%% file: tecs20.tex
\documentclass[acmsmall]{acmart}

\AtBeginDocument{%
  \providecommand\BibTeX{{%
    \normalfont B\kern-0.5em{\scshape i\kern-0.25em b}\kern-0.8em\TeX}}}

\setcopyright{acmcopyright}
\copyrightyear{2020}
\acmYear{2020}
\acmDOI{10.1145/1122445.1122456}

\acmJournal{TECS}
\acmVolume{0}
\acmNumber{0}
\acmArticle{0}
\acmMonth{1}

\usepackage[redeflists]{IEEEtrantools}
\DeclareMathAlphabet{\mymathbb}{U}{BOONDOX-ds}{m}{n}
\newcommand{\tech}{\text{{DFSynthesizer}}}{}
\newcommand{\sm}{\text{{SpiNeMap}}}{}
\newcommand{\pc}{\text{{PyCARL}}}{}
\newcommand{\ineq}[1]{\footnotesize$#1$\normalsize}{}
\usepackage[linesnumbered,ruled]{algorithm2e}
\usepackage{threeparttable}
\usepackage{amsmath}
\newtheorem{Tlemma}{Lemma}
\newenvironment{Lemma}[1]
{\begin{Tlemma}\noindent\textsc{}\itshape}
{\end{Tlemma}}
\newtheorem{Tdef}{Definition}
\newenvironment{Definition}[1]
    {\begin{Tdef}\noindent\textsc{(#1)}\itshape}
    {\end{Tdef}}
\newcommand{\actor}[1]{\textbf{\emph{{#1}}}}
\usepackage{multirow}
\usepackage{xcolor}
\usepackage{subfig}

\SetCommentSty{mycommfont}    
\usepackage{pifont}
\let\oldding\ding
\renewcommand{\ding}[2][1]{\scalebox{#1}{\oldding{#2}}}
\newcommand{\mr}[1]{\textcolor{black}{#1}}
\newcommand{\minor}[1]{\textcolor{black}{#1}}
\newcommand{\bigO}{\mathcal{O}}
\begin{document}
\bstctlcite{IEEEexample:BSTcontrol}
\title{DFSynthesizer: Dataflow-based Synthesis of Spiking Neural Networks to Neuromorphic Hardware}


\author{Shihao Song}
\author{Harry Chong}
%\author{Jacob Baron}
\author{Adarsha Balaji}
\author{Anup Das}
\author{James Shackleford}
\author{Nagarajan Kandasamy}
\email{anup.das@drexel.edu}
\affiliation{%
  \institution{Drexel University}
  \streetaddress{3141 Chestnut Street}
  \city{Philadelphia}
  \state{PA}
  \country{USA}
  \postcode{19104}
}

% \affiliation{%

% \affiliation{%

% \affiliation{%

% \affiliation{%

% \affiliation{%

% \affiliation{%

% \affiliation{%

\renewcommand{\shortauthors}{Song, et al.}

\begin{abstract}
  \input{sections/abstract}
\end{abstract}




%\setcopyright{acmcopyright}
%\acmJournal{TECS}
%\acmYear{2021} \acmVolume{1} \acmNumber{1} \acmArticle{1} \acmMonth{1} \acmPrice{15.00}\acmDOI{10.1145/3479156}

\maketitle

\section{Introduction}\label{sec:introduction}
\input{sections/introduction}

\minor{
\section{Scope and High-Level Overview of \tech{}}\label{sec:high_level}
}

\input{sections/overview}

\section{Program Analysis and Workload Generation}\label{sec:formatting}
\input{sections/formatting}

\section{Program Compilation and Performance Estimation}\label{sec:compilation}
\input{sections/compiler}

\section{Resource Allocation and Hardware Mapping}\label{sec:resource_allocation}
\input{sections/allocation}

\section{Scheduling and Performance Guarantee}\label{sec:scheduling}
\input{sections/scheduling}

\section{Evaluation Methodology}\label{sec:evaluation}
\input{sections/evaluation}

\section{Results and Discussions}\label{sec:results}
\input{sections/results}

\section{Related Works}\label{sec:realted_works}
\input{sections/related_works}

\section{Conclusions}\label{sec:conclusions}
\input{sections/conclusions}

\begin{acks}
This work is supported by 1) the National Science Foundation Award CCF-1937419 (RTML: Small: Design of System Software to Facilitate Real-Time Neuromorphic Computing) and 2) the National Science Foundation Faculty Early Career Development Award CCF-1942697 (CAREER: Facilitating Dependable Neuromorphic Computing: Vision, Architecture, and Impact on Programmability).
\end{acks}

%\balance
\bibliographystyle{IEEEtranSN}
\bibliography{commands,disco,external}

\appendix

\section{Converting Analog Operations to Spiking Equivalent}
\begin{figure}[h!]
	\centering
	\vspace{-5pt}
	\centerline{\includegraphics[width=0.99\columnwidth]{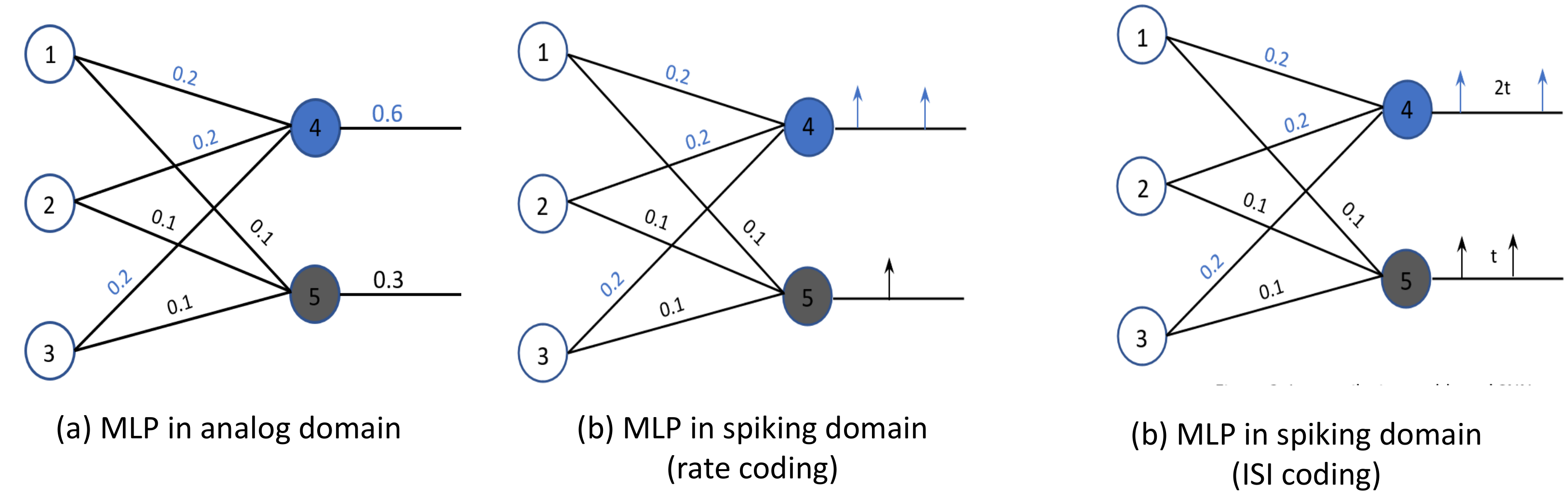}}
	\vspace{-10pt}
	\caption{Example of converting an analog MLP to its spiking equivalent.}
	\vspace{-10pt}
	\label{fig:mlp_example}
\end{figure}

\minor{
In this section, we briefly elaborate how an analog operation such as Rectified Linear Unit (ReLU) is implemented using Spiking Neural Network (SNN). The output \ineq{Y} of a ReLU activation function is given by
\begin{equation}
    \label{eq:relu_fn}
    \footnotesize Y = \max{0,\sum_i w_i*x_i},
\end{equation}
where \ineq{w_i} is the weight and \ineq{x_i} is the activation on the \ineq{i^\text{th}} synapse of the neuron. To map the ReLU activation function, we consider a particular type of spiking neuron model known as an Integrate and Fire (IF) neuron model. The IF spiking neuron's transfer function can be represented as
\begin{equation}
    \label{eq:if_neuron}
    \footnotesize v_m(t+1) = v_m(t) + \sum_i w_i*x_i(t),
\end{equation}
where \ineq{v_m(t)} is the membrane potential of the IF neuron at time \ineq{t}, \ineq{w_i} is the weight, and \ineq{x_i(t)} is the activation on the \ineq{i^\text{th}} synapse of the neuron at time \ineq{t}. The IF spiking neuron integrates incoming spikes (\ineq{X_i}) and generates an output spike (\ineq{Y_\text{spike}}) when the membrane potential (\ineq{v_m}) exceeds the threshold voltage (\ineq{v_\text{th}}) of the IF neuron. Therefore, by ensuring that the output spiking rate \ineq{Y_\text{spike}} is proportional to the ReLU activation \ineq{Y}, i.e., \ineq{Y_\text{spike}\propto Y}, we accurately convert the ReLU activation to the spike-based model.
To further illustrate this, we consider the multi-layer perceptron (MLP) of Figure~\ref{fig:mlp_example}a and its SNN conversion using rate-based encoding (Figure~\ref{fig:mlp_example}b) and inter-spike interval (ISI) encoding (Figure~\ref{fig:mlp_example}c).
}

\minor{
In Figure~\ref{fig:mlp_example}a, neurons 1,2 and 3 are the input neurons and neurons 4 and 5 are the output neurons. To keep the model simple, let us consider the case where the activations of the input neurons 1,2 and 3 are equal to 1. Using Equation~\ref{eq:relu_fn}, we know that the output of neurons 4 and 5 are 0.6 and 0.3, respectively. 
Figures~\ref{fig:mlp_example}b and~\ref{fig:mlp_example}c show the mapped SNN model, using rate-based and inter-spike interval encoding schemes, respectively. In the rate-based model in Figure~\ref{fig:mlp_example}b, the rate of spikes generated is expected to be proportional to the output of neurons 4 and 5 in the MLP. In the case of the ISI-based SNN model, the inter-spike interval of the spikes generated by neurons 4 and 5 is expected to be proportional to the output generated in the MLP, as shown in Figure~\ref{fig:mlp_example}c.
}

\minor{
We note that non-linear activation functions such as sigmoid and tanh cannot be accurately mapped to a spike-based model. This can be attributed to the transfer function of a biological spiking neuron (neuron response curve) closely resembling a ReLU and not sigmoid and tanh activation functions. 
While approximate implementations of the sigmoid and tanh operators using spiking neurons can be found in literature, they induce significant inaccuracies into the conversion process and require more resources (neurons) to implement. 
The tanh activation function, for instance, generates output values ranging between -1.0 to 1.0. In order to represent the tanh function in a spike-based model, both excitatory and inhibitory spiking neurons will be required to represent the positive and negative output values, respectively. This will require doubling the number of spiking neurons needed to represent the tanh activation function.
}








\end{document}

%% file: sections/abstract.tex
Spiking Neural Networks (SNN) are an emerging computation model, which uses event-driven activation and bio-inspired learning algorithms. SNN-based machine-learning programs are typically executed on tile-based neuromorphic hardware platforms, where each tile consists of a computation unit called crossbar, which maps neurons and synapses of the program. 
%comprising the VLSI implementation of biological neurons and synapses. 
However, synthesizing such programs on an off-the-shelf neuromorphic hardware is challenging. This is because of the inherent resource and latency limitations of the hardware, which impact both model performance, e.g., accuracy, and hardware performance, e.g., throughput.
We propose \tech{}, an end-to-end framework for synthesizing SNN-based machine learning programs to neuromorphic hardware. 
The proposed framework works in four steps. First, it analyzes a machine-learning program and generates SNN workload using representative data.
%from an abstract representation of a machine learning model. 
Second, it partitions the SNN workload and generates clusters that fit on crossbars of the target neuromorphic hardware.
%, minimizing key hardware metrics such as resource utilization, energy, and latency. 
Third, it exploits the rich semantics of Synchronous Dataflow Graph (SDFG) to represent a clustered SNN program, allowing for performance analysis in terms of key hardware constraints such as number of crossbars, dimension of each crossbar, buffer space on tiles, and tile communication bandwidth.
%allocates resources to the clusters, allowing performance estimation. 
Finally, it uses a novel scheduling algorithm to execute clusters on crossbars of the hardware, guaranteeing hardware performance. 
%\tech{} exploits the rich semantics of dataflow representation, allowing it analyze a system in terms of key properties such as throughput requirement and buffer availability.
We evaluate \tech{} with 10 commonly used machine-learning programs. Our results demonstrate that \tech{} provides much tighter performance guarantee compared to 
%state-of-the-art approaches.
current mapping approaches.

%% file: sections/introduction.tex
\begin{figure}[h!]
	\centering
	%\vspace{-5pt}
	\centerline{\includegraphics[width=0.79\columnwidth]{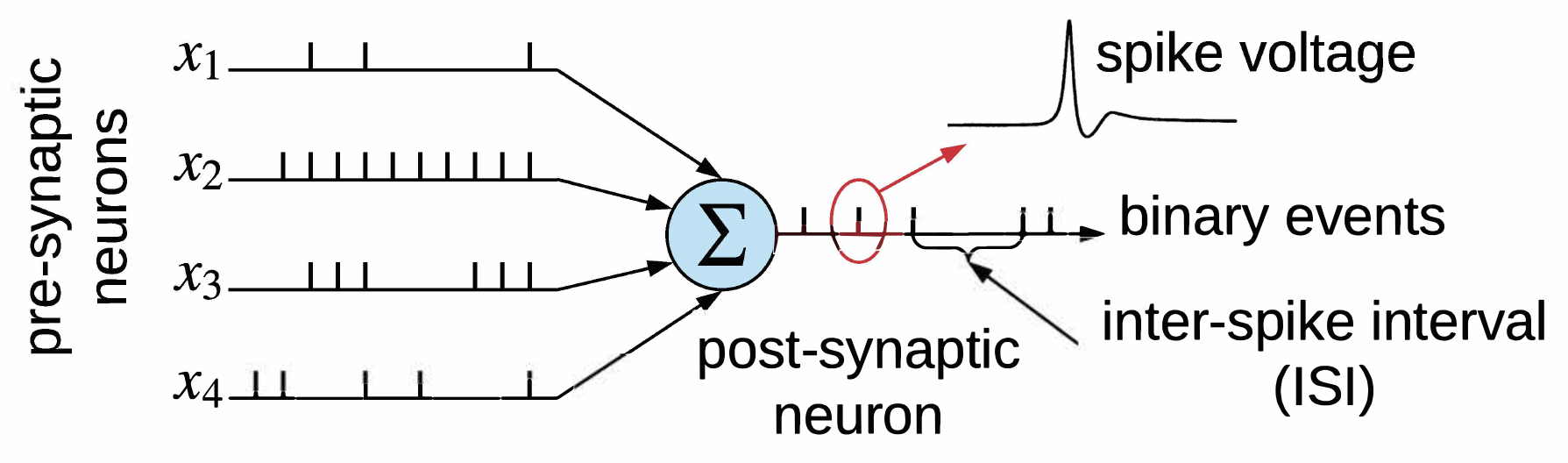}}
	%\vspace{-10pt}
	%\caption{An example of spiking neural network.}
	\caption{Integration of spike trains at the post-synaptic neuron from four pre-synaptic neurons in a Spiking Neural Network (SNN). Each spike is a voltage waveform of time duration to the order of ms.}
	%\vspace{-10pt}
	\label{fig:snn}
\end{figure}

\mr{Spiking Neural Network (SNN) is an emerging computing model that uses spike-based computations and bio-inspired learning algorithms~\cite{maass1997networks}.}
In an SNN, pre-synaptic neurons communicate information encoded in spike trains to post-synaptic neurons, via synapses (see Fig.~\ref{fig:snn}). 
Performance, e.g., accuracy of an SNN model, is assessed in terms of the inter-spike interval (ISI), which is defined as inverse of the mean firing rate of the neurons.

\begin{figure}[h!]
	\centering
	%\vspace{-5pt}
	\centerline{\includegraphics[width=0.99\columnwidth]{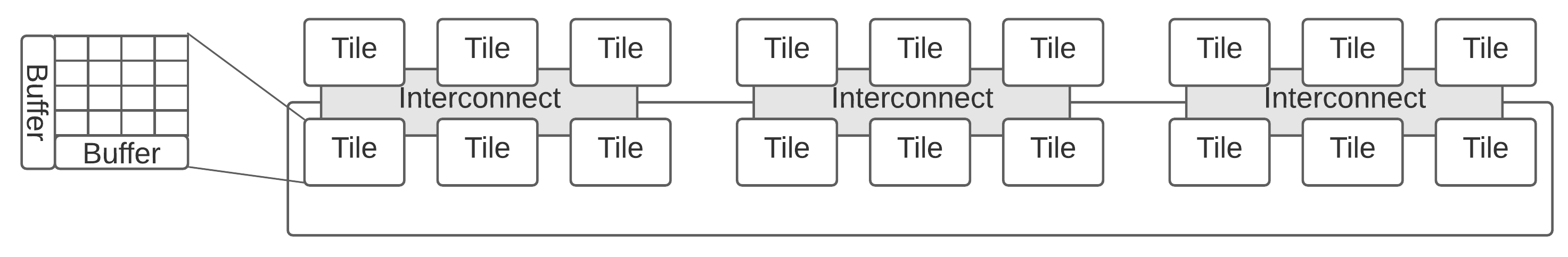}}
	%\vspace{-10pt}
	%\caption{An example of spiking neural network.}
	\caption{A tile-based neuromorphic architecture~\cite{catthoor2018very}, which is representative of many neuromorphic platforms such as DYNAP-SE~\cite{dynapse}, TrueNorth~\cite{truenorth}, and Loihi~\cite{loihi}.}
	%\vspace{-10pt}
	\label{fig:tile}
\end{figure}

SNNs are typically executed on neuromorphic hardware platforms such as DYNAP-SE~\cite{dynapse}, TrueNorth~\cite{truenorth}, and Loihi~\cite{loihi}. These hardware platforms are designed as a tile-based architecture with a shared, hierarchical interconnect to facilitate inter-tile communication (see Fig.~\ref{fig:tile})~\cite{catthoor2018very}. Each tile consists of a crossbar for mapping neurons and synapses, and input and output buffer space for communicating spikes over the interconnect. A crossbar is a 2D organization of horizontal and vertical wires, where the horizontal wires are connected to pre-synaptic neurons while the vertical wires are connected to post-synaptic neurons. Non-Volatile Memory (NVM) cells are placed at the crosspoints of each crossbar to implement storage of synaptic weights~\cite{mallik2017design,Burr2017}.\footnote{Beyond neuromorphic computing, NVMs are also used as main memory for conventional computing using shared-memory computers~\cite{hebe,palp,shihao_igsc,mneme,datacon}.}

Energy consumed by neuromorphic hardware can be several orders of magnitude lower than a conventional machine-learning accelerator such as Eyeriss~\cite{eyeriss}. This is due to low-power VLSI implementation of analog neurons~\cite{indiveri2003low}, low-power and high-density NVM-based synaptic storage~\cite{Burr2017}, as well as distributed computing and storage architecture using crossbars. Given these advantages, a neuromorphic hardware can implement machine-learning tasks for power-constrained platforms such as embedded systems and edge nodes of the Internet-of-Things (IoT)~\cite{atzori2010internet}.

Unlike conventional von-Neumann computing systems, where CPUs compute by exchanging data centrally from the main memory, synthesizing, i.e., compiling and mapping a machine-learning program on a neuromorphic hardware is challenging. This is because in a neuromorphic hardware, computation units (i.e., the neurons) and storage units (i.e., the synapses) are distributed within the hardware as crossbars. It is therefore important to properly partition a large SNN model such that it can be mapped efficiently to the underlying resources. Additionally, each crossbar also presents limitations on how many pre-synaptic connections are allowed per post-synaptic neuron, and how much buffer space is available to send and receive spikes over the interconnect. These hardware limitations impact both model accuracy and hardware performance such as throughput, latency, and energy consumption.

We develop \textbf{\tech{}}, a systematic and end-to-end framework to analyze and map machine-learning programs to state-of-the-art neuromorphic hardware, while guaranteeing performance. Following are our key \textbf{contributions}.\footnote{\mr{Contributions 2, 3, and 4 appeared in our prior work \cite{dfsynthesizer}. This work introduces the contributions 1, 5, and 6.}}

\begin{itemize}
    \item \textbf{Contribution 1.} We present an approach to analyze machine-learning programs and generate SNN workload using representative data. Our framework allows workload generation with only a modest impact on model performance.
    \item \textbf{Contribution 2.} We present an approach to decompose and partition complex SNN workloads and generate clusters of neurons and synapses such that each cluster can fit onto the resources of a crossbar in the hardware.
    \item \textbf{Contribution 3.} We exploit the rich semantics of Synchronous Dataflow Graphs (SDFGs)~\cite{lee1987synchronous} to represent clustered SNN programs. This allows for the SNN's performance, e.g., throughput, to be estimated on the hardware as a function of key properties such as number of crossbars, dimension of crossbars, buffer space on tiles, and tile communication bandwidth.
    \item \textbf{Contribution 4.} We develop a novel scheduling algorithm based on Self-Timed Execution for executing clusters on crossbars of a neuromorphic hardware, providing performance guarantee in scenarios with dynamic resource availability. 
    \item \textbf{Contribution 5.} We propose a design-space exploration framework incorporating \tech{} that allows the Pareto-space of different SNN mappings to hardware to be explored while considering other hardware metrics such as energy, latency, and reliability.
    \item \textbf{Contribution 6.} We evaluate \tech{} using 10 machine learning programs that are representative of the three most commonly used neural network classes --- convolutional neural network (CNN), multi-layer perceptron (MLP), and recurrent neural network (RNN).
\end{itemize}

%% file: sections/overview.tex
\minor{
\tech{} is developed for supervised machine learning approaches, where a machine-learning model is first trained using representative data from the field. Machine learning inference refers to generating output from the trained model by feeding live data. 
To improve energy efficiency, the inference is performed on a neuromorphic hardware. 
Once deployed on the hardware, the model is expected to perform inference in real-time on a continuous basis from data collected using sensors.\footnote{\minor{Camera sensors are used for image classification models, e.g., LeNet, AlexNet, and VGG16, while electrocardiogram sensors are used for heart-rate classification and estimation models. See our evaluation setup in Section~\ref{sec:evaluation}.}}
 Therefore, a key performance metric for neuromorphic hardware performing real-time inference is throughput, defined as the number of frames processed per unit time, where a frame is defined as an individual image (for image-based models) or a window of time-series data.\footnote{\minor{By maximizing the throughput, \tech{} minimizes the time to process individual frame using the neuromorphic inference hardware, which makes \tech{} applicable to both real-time and non real-time applications.}}
}

\begin{figure}[h!]
	\centering
	%\vspace{-5pt}
	\centerline{\includegraphics[width=0.99\columnwidth]{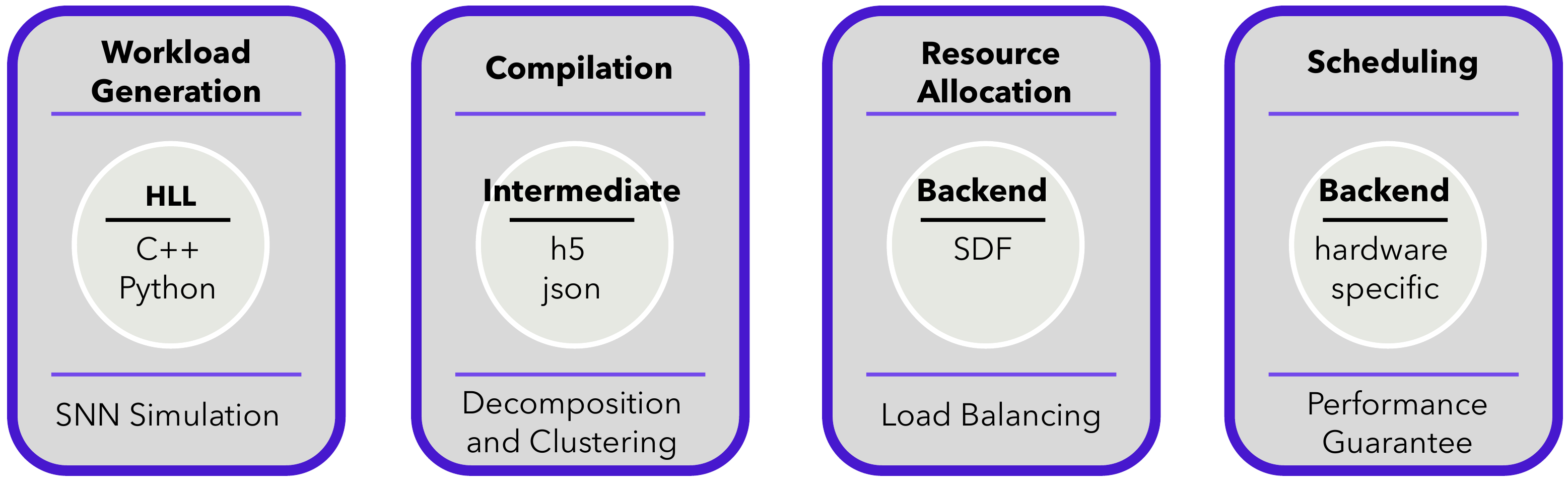}}
	%\vspace{-10pt}
	%\caption{An example of spiking neural network.}
	\caption{High-level overview of \tech{}. A machine learning program is analyzed and mapped to the hardware using the proposed 4-step methodology.}
	%\vspace{-10pt}
	\label{fig:framework}
\end{figure}

Figure~\ref{fig:framework} illustrates the proposed end-to-end framework of \tech{}, which synthesizes, i.e., compiles and maps a machine learning program to a neuromorphic hardware in four steps. First, it analyzes a machine learning program written in a high-level language such as Python and C/C++ to generate SNN workload (Section~\ref{sec:formatting}).  Second, it compiles SNN workloads to an intermediate representation format (h5 and json), performing spatial decomposition and clustering to fit onto the resources of a crossbar (Section~\ref{sec:compilation}). Third, it uses Synchronous Dataflow Graph (SDF) to represent clustered SNN (in XML representation), allocating resources to the clusters considering hardware resource constraints (Section~\ref{sec:resource_allocation}). Finally, it schedules the SDF representation of a clustered SNN to the hardware crossbars, guaranteeing performance (Section~\ref{sec:scheduling}). 

%Overall, \tech{} provides performance guarantee for executing machine learning programs on neurromorphic hardware.

%% file: sections/formatting.tex
In this step, a machine-learning program is analyzed to generate its workload. In the following, we discuss the steps involved in the workload generation.

\subsection{\mr{Workflow for Workload Generation}}
\begin{figure}[h!]
	\centering
	%\vspace{-5pt}
	\centerline{\includegraphics[width=0.99\columnwidth]{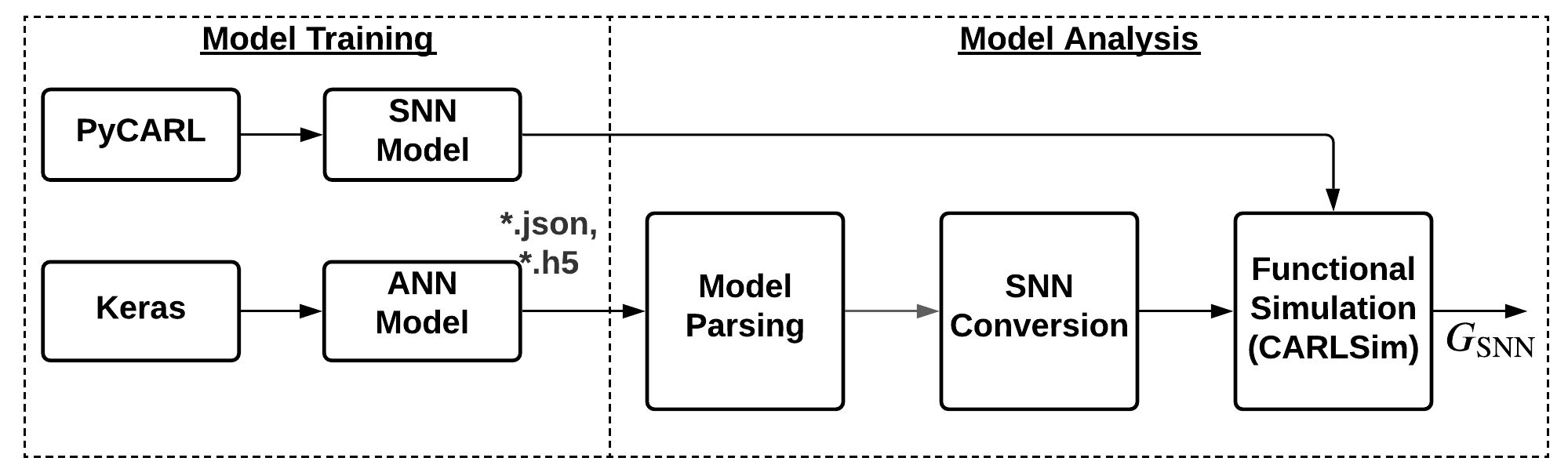}}
	%\vspace{-10pt}
	%\caption{An example of spiking neural network.}
	\caption{Workflow of the workload generation step of \tech{}.}
	%\vspace{-10pt}
	\label{fig:workflow_workload_gen}
\end{figure}

\mr{
Figure~\ref{fig:workflow_workload_gen} summarizes the workflow of the workload generation step of \tech{}, where a machine-learning program is analyzed to generate its workload which is then used to map the application to a neuromorphic hardware.
}

\mr{
\tech{} can incorporate both Artificial Neural Networks (ANNs) and Spiking Neural Networks (SNNs) in its workflow. 
At a high level, the proposed workflow consists of a model training component followed by model analysis.
In the following, we elaborate on these components. %of this workflow.
}

\subsection{\mr{Model Training}}
\subsubsection{\mr{\underline{Training Artificial Neural Networks}}}
\mr{
\tech{}'s frontend is integrated with Keras~\cite{keras}, which is used to define a model and train it on a database. Keras utilizes Tensorflow backend~\cite{tensorflow}. \tech{} also supports other frameworks such as PyTorch~\cite{pytorch}. To demonstrate the capabilities of \tech{}, we evaluate it with three Convolutional Neural Network (CNN) architectures -- 1) LeNet~\cite{lenet}, trained on MNIST handwritten digit dataset~\cite{mnist}, 2) AlexNet~\cite{alexnet}, trained on ImageNet dataset~\cite{imagenet}, and 3) VGGNet~\cite{vggnet}, trained on ImageNet dataset. These models are derived from the MLPerf~\cite{mlperf} dataset and instantiated in Keras. We use a Lambda workstation with two GPUs (see our evaluation setup in Section~\ref{sec:evaluation}) to train these models. 
}

\subsubsection{\mr{\underline{Training Spiking Neural Networks}}}
\mr{\tech{}'s frontend supports training SNN models using PyCARL~\cite{pycarl}, a Python frontend to CARLsim~\cite{carlsim}.
%PyNN~\cite{pynn}. 
CARLsim facilitates SNN simulations using CPUs and multi-GPUs.
PyCARL is designed to integrate with PyNN~\cite{pynn}, which provides a common frontend to different SNN simulators with various degrees of neurobiological details. We use CARLsim for model training. CARLsim's support for built-in biologically realistic neuron, synapse, current and emerging learning models and continuous
integration and testing, make it an easy-to-use and powerful simulator of biologically-plausible SNN models. \tech{} can also utilize other SNN simulators such as Brian~\cite{brian}, NEST~\cite{nest}, and NEURON~\cite{neuron} for model training.
}

\subsection{\mr{Model Analysis}}
\subsubsection{\mr{\underline{Model Parsing and Conversion}}}
\mr{
Unfortunately, ANN models cannot be executed directly on event-driven neuromorphic hardware platforms such as DYNAP-SE~\cite{dynapse}, TrueNorth~\cite{truenorth}, and Loihi~\cite{loihi}. Recently, many tools have been proposed to convert ANN operations to SNNs. Examples include Nengo~\cite{nengo}, N2D2~\cite{n2d2}, and SNNToolBox~\cite{zurich_converter}. A common limitation of these toolboxes is that they are open-loop converters, meaning that the conversion is performed considering performance degradation only. In our prior work~\cite{jolpe18}, we have proposed a closed-loop conversion mechanism, where the conversion of analog operations to spiking equivalent is performed considering the energy consumption on hardware. These conversion steps are briefly discussed below.\footnote{\mr{The conversion framework was introduced in~\cite{jolpe18} for converting CNN-based HeartClass application to its equivalent SNN representation. We used this application to evaluate \tech{}. Additionally, we have extended the conversion framework to add other key functionalities such as Layer Flattening, Concatenation, Binary Weight Activation, and Non-Zero Biases. These new functionalities allowed the conversion framework to convert state-of-the-art CNN architectures such as LeNet, AlexNet, and VGG16, which are used to evaluate \tech{}.}}
}

\begin{enumerate}
    \item \emph{ReLU Activation Functions:} This is implemented as the approximate firing rate of a leaky integrate and fire (LIF) neuron.
    
    \item \emph{Bias:} A bias is represented as a constant input current to a neuron, the value of which is proportional to the bias of the neuron in the corresponding analog model.
    
    \item \emph{Weight Normalization:} This is achieved by setting a factor \ineq{\lambda} to control the firing rate of spiking neurons.
    
    \item \emph{Softmax:} To implement softmax, an external Poisson spike generator is used to generate spikes proportional to the weighted sum accumulated at each neuron.
    
    \item \emph{Max and Average Pooling:} To implement max pooling, the neuron which fires first is considered to be the winning neuron, and therefore, its responses are forwarded to the next layer, suppressing the responses from other neurons in the pooling function. To implement average pooling, the average firing rate (obtained from total spike count) of the pooling neurons are forwarded to the next layer.
\end{enumerate}

\mr{
%In order to use the conversion framework to convert CNN architectures such as LeNet, AlexNet, and VGGNet, we have extended the framework to support the following new functionalities.
We have extended our framework with the following new functionalities to allow for the conversion of CNN architectures such as LeNet, AlexNet, and VGGNet to their spiking counterparts.
\begin{enumerate}
    \item \emph{1-D Convolution:} The 1-D convolution is implemented to extract patterns from inputs in a single spatial dimension. A 1xn filter, called a kernel, slides over the input while computing the element-wise dot-product between the input and the kernel at each step.
    \item \emph{Residual Connections:} Residual connections are implemented to convert the residual block used in CNN models such as ResNet. Typically, the residual connection connects the input of the residual block directly to the output neurons of the block, with a synaptic weight of `1'. This allows for the input to be directly propagated to the output of the residual block while skipping the operations performed within the block.
    \item \emph{Flattening:} The flatten operation converts the 2-D output of the final pooling operation into a 1-D array. This allows for the output of the pooling operation to be fed as individual features into the decision-making fully connected layers of the CNN model.
    \item \emph{Concatenation:} The concatenation operation, also known as a merging operation, is used as a channel-wise integration of the features extracted from 2 or more layers into a single output.
\end{enumerate}
}

Table~\ref{tab:conversion_accuracy} reports the accuracy impact due to the SNN conversion of three state-of-the-art supervised CNN models. 
 \mr{
 These accuracy numbers are obtained from CARLsim~\cite{carlsim}, which allows functional simulation and performance estimation of SNN-based applications.
 }
 We use these three converted CNN models to evaluate \tech{} (See Section~\ref{sec:evaluation}).

\begin{table}[h!]
	\renewcommand{\arraystretch}{1.4}
	\setlength{\tabcolsep}{7pt}
	\centering
	\begin{threeparttable}
	{\fontsize{8}{10}\selectfont
	    %\vspace{-10pt}
		\begin{tabular}{c|c|c|c|c|c|c|c|c}
			\hline
			\multirow{2}{*}{\textbf{Application}} & \multicolumn{2}{|c|}{\minor{Top-1 Accuracy (\%)}} & \multirow{2}{*}{\textbf{Application}} & \multicolumn{2}{|c|}{\minor{Top-1 Accuracy (\%)}} & \multirow{2}{*}{\textbf{Application}} & \multicolumn{2}{|c}{\minor{Top-1 Accuracy (\%)}}\\
			\cline{2-3}\cline{5-6}\cline{8-9}
			& Original & SNN & & Original & SNN & & Original & SNN\\
			\hline
			LeNet & 94.98\% & 94.08\% & AlexNet & 74.1\% & 71.7\% & VGG16 & 93.56\% & 91.62\%\\
			\hline
	\end{tabular}}
	%}
	\end{threeparttable}
	%\vspace{12pt}
	\caption{Accuracy impact due to conversion of three state-of-the-art CNN models to their SNN equivalent. \mr{The original accuracy numbers are obtained by simulating these architectures in Keras~\cite{keras} with Tensorflow backend~\cite{tensorflow}. The converted accuracy numbers reported in the columns marked ``SNN'' are obtained from CARLsim~\cite{carlsim}. We use a multi-GPU machine to simulate these architectures using both Keras and CARLsim. See our evaluation framework in Section~\ref{sec:evaluation}.}}
	\label{tab:conversion_accuracy}
	%\vspace{-10pt}
\end{table}

% \mr{
% In order to parse ANN models, we have build a parser that takes an ANN model in h5 and json formats, and feed it to the converter for conversion to SNN.
% }

\subsubsection{\mr{\underline{Workload Generation}}}
\mr{
The SNN model (or the converted ANN model) is analyzed in CARLsim to generate the following information.
}

\begin{itemize}
    \item \emph{\textcolor{black}{Spike Data:}} the exact spike times of all neurons in the SNN model. We let \ineq{spk(i)} represents a list of spike times of the \ineq{i^\text{th}} neuron in the model.
    \item \emph{\textcolor{black}{Weight Data:}} the synaptic strength of all synapses in the SNN model. We let \ineq{w(i,j)} represents the synaptic weight of the connection between the \ineq{i^\text{th}} and \ineq{j^\text{th}} neurons in the SNN model.
\end{itemize}

The spike and weight data of a trained SNN form the \textbf{SNN workload}.
 Formally, an SNN workload is defined as 
\begin{Definition}{SNN Workload}
{An SNN Workload ${\mathbf{G_{SNN} = (N,S,W)}}$ is a directed graph consisting of a finite set \ineq{N} of neurons, a set \ineq{S} of spikes, and a set \ineq{W} of synapses between the neurons.}
%$\tau \in \mathbb{N}\setminus\{0\}$ representing the execution time of a ($\tau(a)$).}
\end{Definition}

%% file: sections/compiler.tex
In this step, \tech{} clusters a given machine-learning model to map onto the crossbars of a neuromorphic hardware. 
\mr{
To do so, we first introduce the system architecture and then discuss the clustering step needed to map applications to this architecture.
}

\subsection{System Architecture}
\begin{figure}[h!]
	\centering
	%\vspace{-5pt}
	\centerline{\includegraphics[width=0.99\columnwidth]{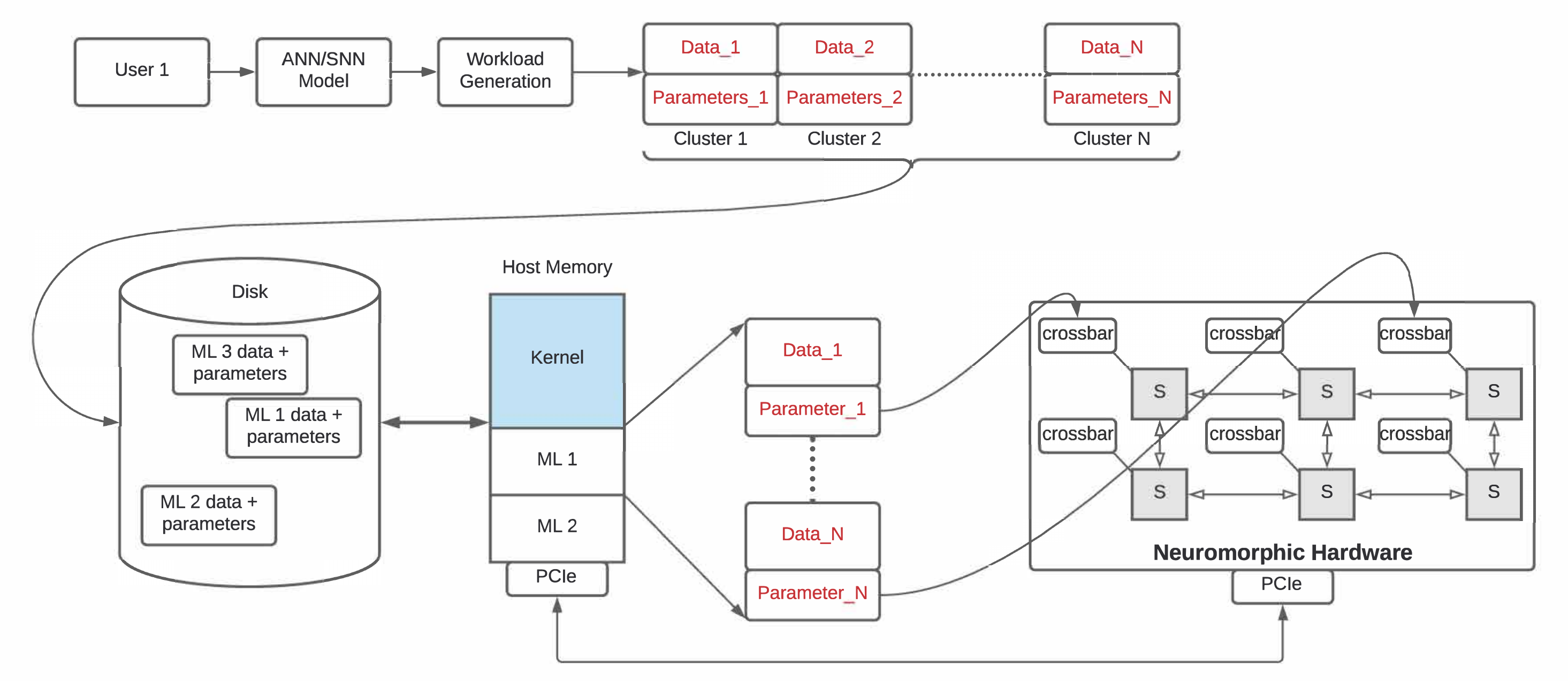}}
	%\vspace{-10pt}
	%\caption{An example of spiking neural network.}
	\caption{\minor{Our system architecture, integrating a neuromorphic hardware. \tech{} is designed for crossbar-based neuromorphic hardware~\cite{catthoor2018very,gopalakrishnan2020hfnet,ankit2017trannsformer,hu2016dot}. This is representative of many recent neuromorphic designs. To evaluate \tech{}, we have configured our evaluation setup to model the DYNAP-SE hardware~\cite{dynapse}.}}
	%\vspace{-10pt}
	\label{fig:system_architecture}
\end{figure}

\mr{
Figure~\ref{fig:system_architecture} illustrates our system architecture.
}
\minor{
\tech{} is designed for crossbar-based neuromorphic hardware designs as shown in Figure~\ref{fig:tile}. This is representative of many recent neuromorphic designs~\cite{catthoor2018very,gopalakrishnan2020hfnet,ankit2017trannsformer,hu2016dot}.   
}
\mr{
A machine learning model (ANN or SNN) is first analyzed to generate its workload (Section~\ref{sec:formatting}). This workload is then partitioned to generate clusters, where each cluster consists of a fraction of the neurons and synapses of the original machine learning model. The cluster workload is stored in a disk along with other machine learning workloads.
To execute a specific workload on the neuromorphic hardware, it is first loaded into the host memory and then the clusters are programmed on to the crossbars of the hardware via the PCIe interface.\footnote{\mr{Although we illustrate the crossbars to be interconnected in a mesh-based architecture such as Networks-on-Chip (NoC)~\cite{noc_benini}, \tech{} can work with other interconnect types such as Segmented Bus~\cite{balaji2019exploration}.}}
}

\mr{
In the remainder of this section, we describe the workload compilation step of \tech{}, which consists of the following two design components -- Workload Decomposition and Workload Clustering. We conclude this section by providing a dataflow modeling approach for clustered workloads and performance estimation using such model.
}

\subsection{\mr{Workload Decomposition}}
\begin{figure}[h!]
	\centering
	%\vspace{-5pt}
	\centerline{\includegraphics[width=0.69\columnwidth]{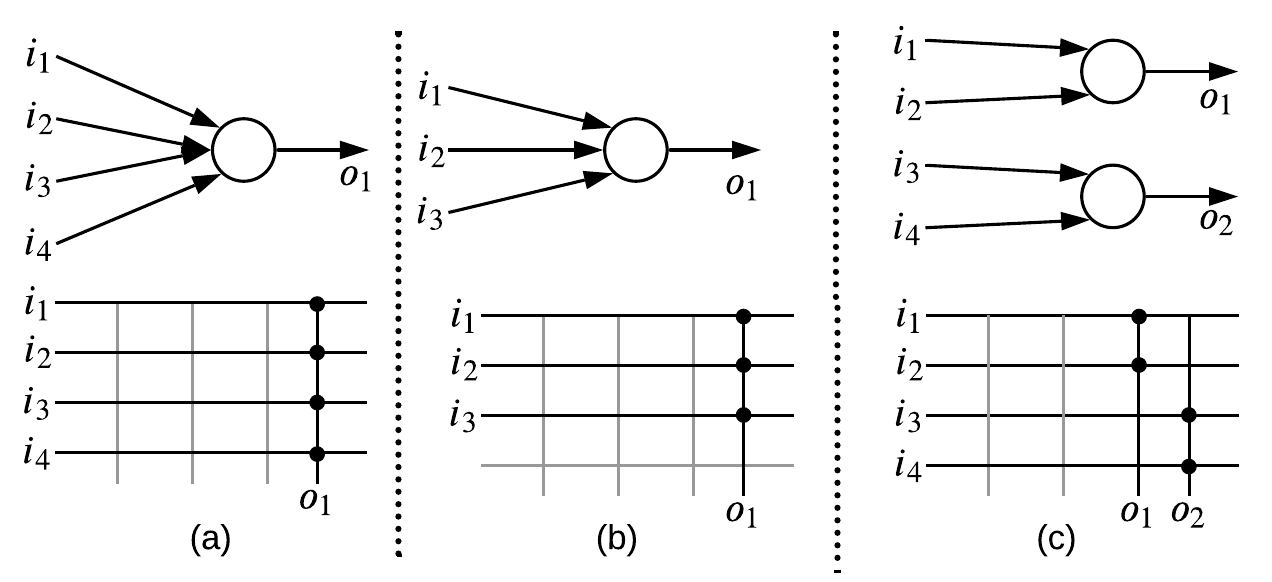}}
	%\vspace{-10pt}
	%\caption{An example of spiking neural network.}
	\caption{Example mapping of a) one 4-input, b) one 3-input, and c) two 2-input neurons on a $4 \times 4$ crossbar.}
	%\vspace{-10pt}
	\label{fig:crossbar_mapping}
\end{figure}

We note that each $N \times N$ crossbar in a neuromorphic hardware can accommodate up to \ineq{N} pre-synaptic connections per post-synaptic neuron, with typical value of \ineq{N} set between 128 (in DYNAP-SE) and 256 (in TrueNorth). 
Figure~\ref{fig:crossbar_mapping} illustrates an example of mapping a) one 4-input, b) one 3-input, and c) two 2-input neurons on a $4 \times 4$ crossbar. Unfortunately, neurons with more than 4 pre-synaptic connections per post-synaptic neuron cannot be mapped to the crossbar. In fact, in many complex machine learning models such as AlexNet and VGG16, the number of pre-synaptic connections per post-synaptic neuron is much higher than 128. Therefore, these neurons cannot be mapped to a $128 \times 128$ crossbar in DYNAP-SE. 

To address the above limitation, we have previously proposed a spatial decomposition technique which exploits the firing principle of LIF neurons, decomposing each neuron with many pre-synaptic connections into a sequence of homogeneous fanin-of-two (FIT) neural units~\cite{esl20}. 

\begin{figure}[h!]
	\centering
	%\vspace{-5pt}
	\centerline{\includegraphics[width=0.99\columnwidth]{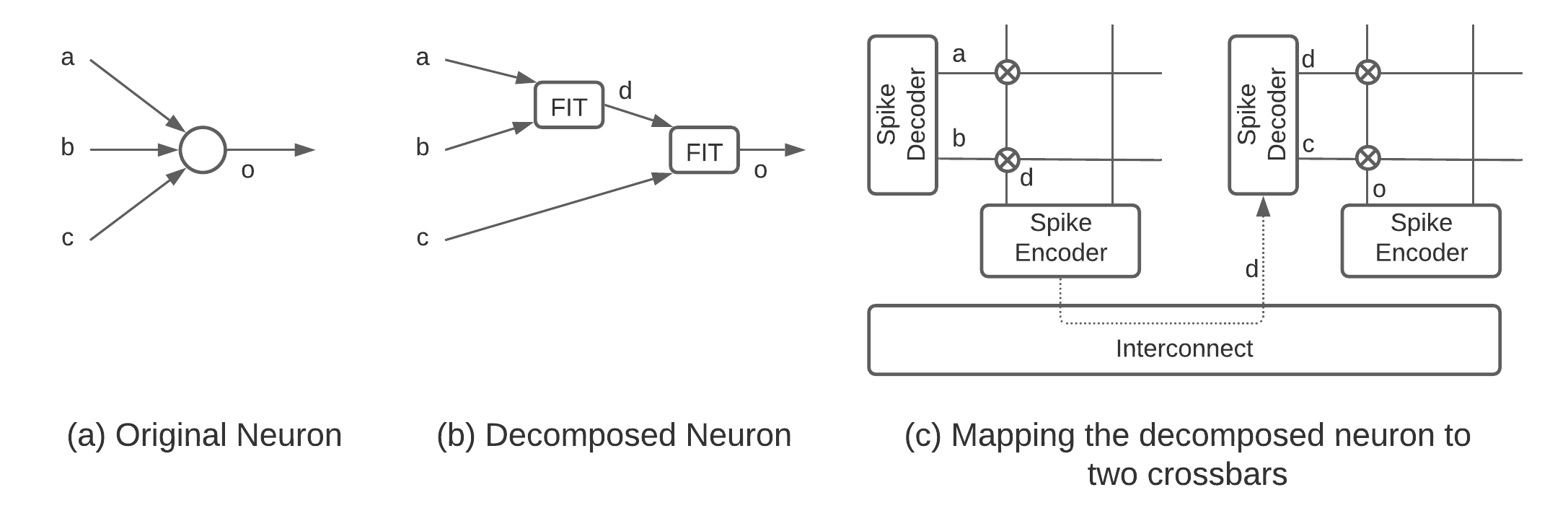}}
	%\vspace{-10pt}
	%\caption{An example of spiking neural network.}
	\caption{\mr{Illustrating the decomposition of a 3-input neuron (a) to a sequence of FIT neural units (b). The mapping of the FIT units to two 2x2 crossbars is shown in (c).}}
	%\vspace{-10pt}
	\label{fig:decomposition_demo}
\end{figure}

\mr{
Figure~\ref{fig:decomposition_demo} illustrates the spatial decomposition using a small example of a 3-input neuron shown in Figure~\ref{fig:decomposition_demo}(a). We consider the mapping of this neuron to 2x2 crossbars. Since each crossbar can accommodate a maximum of two pre-synaptic connections per neuron, the example 3-input neuron cannot be mapped to the crossbar directly. The most common solution is to eliminate a synaptic connection, which may lead to accuracy loss. Figure~\ref{fig:decomposition_demo}(b) illustrates the decomposition mechanism, where the 3-input neuron is implemented using two FIT neural units connected in sequence as shown in Figure~\ref{fig:decomposition_demo}(b). Each FIT unit is similar to a 2-input neuron and it exploits the leaky integrate behavior in hardware to maintain the functional equivalence between Figures~\ref{fig:decomposition_demo}(a) and \ref{fig:decomposition_demo}(b).
}

\mr{
For the sake of completeness, Figure~\ref{fig:decomposition_demo}(c) illustrates the mapping of the decomposed neuron utilizing two 2x2 crossbars. The functionality of the FIT neural units is implemented using the Non-Volatile Memory (NVM) cells of the two crossbars.
}

\mr{
To describe the decomposition Algorithm, we introduce the following notations.
Let \ineq{n_i^1,n_i^2,\cdots,n_i^{m_i}} be the \ineq{m_i} pre-synaptic connections of the neuron \ineq{N_i}. Let \ineq{F_i^1,F_i^2,\cdots,F_i^{m_i-1}} be the (\ineq{m_i-1}) FIT neural units that are generated by spatially decomposing this neuron. The input of unit \ineq{F_i^j} denoted as \ineq{In(F_i^j)} can be represented as 
}

\mr{
\begin{equation}
    \label{eq:spatial_decompose}
    \footnotesize In(F_i^j) = \begin{cases}
    \{n_i^1,n_i^2\} & \text{ for j = 1} \\
    \{n_i^{j+1},Out(F_i^{j-1})\} & \text{ otherwise} 
    \end{cases}~\forall j\in~\{1,2,\cdots,m_i-1\}
\end{equation}
}
\mr{
where \ineq{Out(F_i^j)} is the output of the unit \ineq{F_i^j}. When decomposing a neuron, we note that the first FIT unit uses two of the original inputs of the original neuron. Subsequently, all other FIT units use one of the original inputs and the output of the preceding FIT units as shown in Figure~\ref{fig:decomposition_demo}(b).
}

Formally, a decomposed SNN graph is defined as follows.
\begin{Definition}{Decomposed SNN Graph}
A decomposed SNN graph \ineq{\mathbf{G_{DSNN} = (\textbf{F},\textbf{L})}} is a directed graph consisting of a finite set \ineq{{\textbf{F}}} of FIT neural units and a finite set \ineq{{\textbf{L}}} of links between these units.
%$\tau \in \mathbb{N}\setminus\{0\}$ representing the execution time of a ($\tau(a)$).}
\end{Definition}

Algorithm~\ref{alg:unrolling} shows the pseudo-code of the spatial decomposition technique, which performs the graph transformation \ineq{G_{SNN}\rightarrow G_{DSNN}}. 
\mr{
For each neuron \ineq{N_i} (line 1), a set of inputs to this neuron is obtained (line 2). The first FIT unit is formed using two input inputs (line 3). This is in accordance with Equation~\ref{eq:spatial_decompose} and Figure~\ref{fig:decomposition_demo}(b). The FIT unit is inserted into the decomposed graph \ineq{G_{DSNN}} (line 4). The algorithm then creates the other FIT units iteratively (lines 5-8) using Equation~\ref{eq:spatial_decompose} and stores those units in \ineq{G_{DSNN}}. Finally, the graph \ineq{G_{DSNN}} is returned (line 10). 
}

\mr{
The overall complexity of this algorithm is calculated as follows. The Out for loop (lines 1-9) is executed for the neurons in the original graph \ineq{G_{SNN}}, i.e., for \ineq{|N|} times. Within each iteration, the algorithm creates a total of \ineq{\left(|In(N_i)|-1\right)} FIT units, where \ineq{In(N_i)} is the set of input of neuron \ineq{N_i}. Therefore, the algorithmic complexity is
\begin{equation}
    \footnotesize \text{Complexity} = \bigO\left(\sum_{i=1}^{|N|}\bigg(|In(N_i|-1\bigg)\right) \approx \bigO\left(|W|\right)
\end{equation}
In deriving the final expression, we note that the input connections of all the neurons in the graph \ineq{G_{SNN}} are the edges \ineq{W} in the graph. 
}

\begin{algorithm}[h]
	\scriptsize{
 		\KwIn{\ineq{G_{SNN}= (\textbf{N},\textbf{W})}}
 		\KwOut{\ineq{G_{DSNN}= (\textbf{F},\textbf{L})}}
 		\For(\tcc*[f]{for each node of $G_{SNN}$}){$N_i\in \mathbf{N}$}{
 		    $\{n_i^1,n_i^2,\cdots,n_i^{m_i}\} = \texttt{In}(N_i)$ \tcc*[r]{input links of $N_i$}
 		    Create node $F_i^1$ with $\texttt{In}(F_i^1) = \{n_1,n_2\}$ \tcc*[r]{first FIT unit}
 		    $G_{DSNN}.\texttt{insert}(F_i^1)$\tcc*[r]{insert the FIT neural unit $u_1^i$ in $G_{DSNN}$}
 		    \For(\tcc*[f]{remaining FIT units}){$j=2;j<m_i;j++$}{
 		        Create node $F_i^j$ with $\texttt{in}(F_i^j) = \{n_i^{j+1},F_i^{j-1}\}$\;
 		        $G_{DSNN}.\texttt{insert}(F_i^j)$\;
 		    }
 		}
 		Return $G_{DSNN}$
 	}
	\caption{Spatial decomposition of SNN graph $G_{SNN}$.}
	\label{alg:unrolling}
\end{algorithm}

\subsection{Workload Clustering}
\begin{figure}[h!]
	\centering
	%\vspace{-5pt}
	\centerline{\includegraphics[width=0.99\columnwidth]{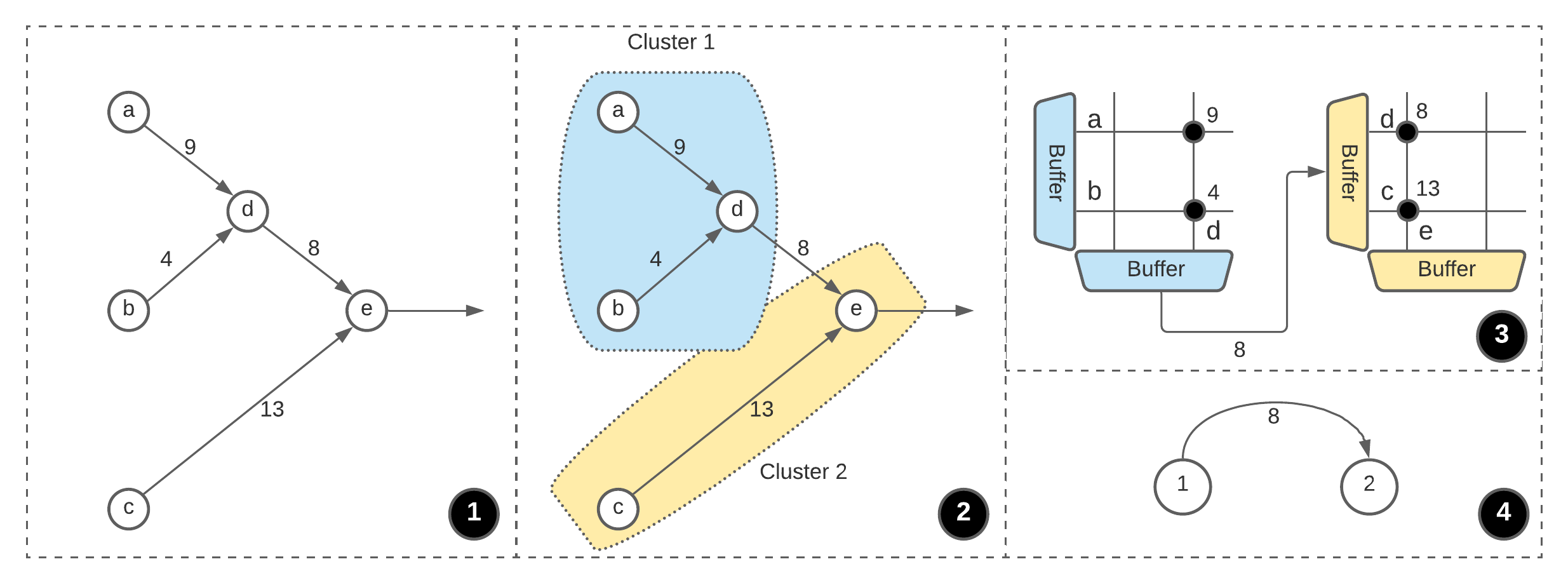}}
	%\vspace{-10pt}
	%\caption{An example of spiking neural network.}
	\caption{Illustration of SNN graph clustering. (\ding{182}) is the original decomposed SNN graph with FIT neural units shown as the nodes and average spikes communicated between them shown on the links. (\ding{183}) shows the partitioning of this graph. (\ding{184}) shows the mapping of the partitions to the two crossbars. (\ding{185}) shows the two clusters generated from the SNN graph of (\ding{182}) considering the constraints of the crossbar.}
	%\vspace{-10pt}
	\label{fig:clustering_demo}
\end{figure}

The decomposed SNN graph is clustered such that each cluster is able to fit onto a crossbar. Figure~\ref{fig:clustering_demo} illustrates the concept using an example of a decomposed SNN graph shown in (\ding{182}). The nodes are the FIT neural units and the links are the synaptic connections. The number on a link represents the average number of spikes communicated between the source and destination FIT units for the representative training data. We consider the mapping of this decomposed SNN graph to a hardware with $2 \times 2$ crossbars. Since a crossbar in this hardware can only accommodate a maximum of 2 pre-synaptic connections, we partition the graph of (\ding{182}) into two partitions (shown in two different colors) in (\ding{183}). These partitions can then be mapped to the two crossbars as shown in (\ding{184}), with an average 8 spikes communicated between the crossbars due to the mapping of the link between neuron \textbf{\textit{d}} and \textbf{\textit{e}} on the shared interconnect of the hardware. Finally, the two clusters generated from the SNN graph are shown in (\ding{185}) along with the inter-cluster communication.

Formally, a clustered SNN graph is defined as follows.
\begin{Definition}{Clustered SNN Graph}
A clustered SNN graph \ineq{\mathbf{G_{CSNN} = (\textbf{A},\textbf{C})}} is a directed graph consisting of a finite set \ineq{{\textbf{A}}} of clusters and a finite set \ineq{{\textbf{C}}} of connections between these clusters.
%$\tau \in \mathbb{N}\setminus\{0\}$ representing the execution time of a ($\tau(a)$).}
\end{Definition}

\mr{Recently, different approaches have been proposed for clustering SNNs. 
Examples include SpiNeMap~\cite{spinemap} for energy minimization and NEUTRAMS~\cite{ji2016neutrams} for performance. See Section~\ref{sec:realted_works} for a comprehensive overview of other state-of-the-art SNN clustering approaches.}
%Balaji et al.~\cite{spinemap} and Das et al.~\cite{psopart} propose a clustering approach to minimize the inter-cluster spike communication. Song et al.~\cite{reneu} and Titirsha et al.~\cite{twisha_thermal,twisha_endurance} propose to minimize the spike activation within each crossbar. Song et al.~\cite{dfsynthesizer} and Balaji et al.~\cite{pycarl,esl20} propose to minimize the number of clusters. 

\mr{We formulate SNN clustering as a graph transformation problem and introduce an efficient algorithm to improve resource utilization. This objective is essential to provide tighter guarantee on performance of SNNs in hardware as we demonstrate in Section~\ref{sec:results}.}%, while optimizing a given objective function. 

The graph transformation \ineq{G_{DSNN}\rightarrow G_{CSNN}} is a classical graph partitioning problem \cite{kernighan1970efficient}, and has been applied in many contexts, including task mapping on multiprocessor systems \cite{das2014communication}. 
%In \cite{dfsynthesizer} for instance, a greedy approach is proposed, roughly based on the Kernighan-Lin Graph Partitioning algorithm~\cite{kernighan1970efficient}, which is shown to be scalable to large SNNs. The objective is to improve the 
%The algorithm sorts the synapses in descending order of the total number of spikes. Next, synapses are packed into clusters, such that the total number of input and output connections of each cluster is less than \ineq{N} for NxN crossbars in the hardware. 
We propose a greedy approach to pack the FIT neural units and synapses of the decomposed SNN graph \ineq{G_{DSNN}} into clusters, improving cluster resource utilization. Algorithm~\ref{alg:clustering} provides the pseudo-code of the clustering algorithm. For each node of the unrolled graph, the algorithm tries to see if the node can be merged into one of the existing clusters (line 3), before creating a new one (lines 4--8). In this algorithm, clusters in \ineq{G_{CSNN}} are sorted in descending order of neuron and synapse utilization (line 12), so that the heavily utilized clusters are first considered for packing neurons and synapses, further improving their utilization.

\begin{algorithm}[h]
	\scriptsize{
		\KwIn{$G_{DSNN} = ({\textbf{F},\textbf{L}})$}
		\KwOut{$G_{CSNN} = (\textbf{A},\textbf{C})$}
		%\texttt{neuron\_list} = sort neurons of the SNN based on their fanin synapses\;
		$G_{CSNN}$ = \{\} and \texttt{cluster\_list} = \{\}\;
		\ForEach{$F_i\in \textbf{F}$}{
		    find $C_j \in \texttt{cluster\_list}$ such that $F_i$ can be packed in $C_j$ while improving neuron and synapse utilization of $C_j$\;
		    \If{$C_j = \emptyset$}{
		        Create new cluster $C_\text{new}$\;
		        Assign $F_i$ and its synaptic connections to $C_\text{new}$\;
		        $G_{CSNN}$.\texttt{push}($C_\text{new}$)\;
		    }\Else{
		        Assign $F_i$ and its synaptic connections to $C_j$\;
		    }
		    sort $G_{CSNN}$ in descending order of neuron and synapse utilizations\;
		}
	}
	\caption{\small Utilization-aware SNN clustering.}
	\label{alg:clustering}
\end{algorithm}

\subsection{Dataflow Modeling of Clustered Workload}
We model a clustered SNN as a Synchronous Data Flow Graph (SDFG) for predictable performance analysis~\cite{lee1987synchronous}. SDFGs are commonly used to model streaming applications that are implemented on a multi-processor system-on-chip~\cite{SB00}. 
%Both pipelined streaming and cyclic dependencies between tasks can be easily modeled in SDFGs. 
These graphs are used to analyze a system in terms of key performance properties such as throughput, execution time, communication bandwidth, and buffer requirements~\cite{Stuijk06dac}. 
Nodes of an SDFG are called \textit{actors}. Each node is a cluster of the clustered SNN graph \ineq{\mathbf{G_{CSNN} = (\textbf{A},\textbf{C})}}. Actors are computed by reading \textit{tokens}, i.e., spikes from their input ports and writing the results of the computation as tokens on the output ports. The number of tokens produced or consumed in one execution of an actor is called the \textit{port rate}. They represent the number of spikes per unit time at the input and output of different clusters in the SNN. Port rates are visualized as annotations on edges. Actor execution is also called \textit{firing}, and it requires a fixed amount of time to execute on a crossbar. Edges in the graph are called \textit{channels} and they represent dependencies among actors.
An actor is said to be {\em ready} when it has sufficient input tokens on all its input channels and sufficient buffer space on all its output channels; an actor can only fire when it is ready.
A set $Ports$ of ports is assumed, and with each port $p \in Ports$, a finite rate $Rate(p) \in \mathbb{N}\setminus\{0\}$ is associated.
Formally, an actor is defined as follows.
\begin{Definition}{Actor}
{An actor $\actor{a}_i$ is a tuple $(I_i,O_i,\tau_i,\mu_i)$ consisting of a set $I_i$ ($\subseteq Ports$) of input ports, a set $O_i$ ($\subseteq Ports$) of output ports with $I_i \cap O_i = \emptyset$, $\tau_i$ is the execution time of $\actor{a}_i$ and $\mu_i$ is its state space, i.e., buffer space needed for communicating spikes on all of its channels.}
\end{Definition}
The source of channel $ch_i^j \in C$ is an output port of actor $\actor{a}_i$, the destination is an input port of actor $\actor{a}_j$. All ports of all actors are connected to precisely one channel, and all channels are connected to ports of some actors. The source and the destination port of channel $ch_i^j$ are denoted by $SrcP(ch_i^j)$ and $DstP(ch_i^j)$ respectively. Channels connected to the input and output ports of an actor $\actor{a}_i$ are denoted by $InC(\actor{a}_i)$ and $OutC(\actor{a}_i$) respectively.

Before an actor $\actor{a}_i$ starts its firing, it requires $Rate(q_i)$ tokens from all $(p,q_i)\in InC(\actor{a}_i)$. When the actor completes execution, it produces $Rate(p_i)$ tokens on every $(p_i,q) \in OutC(\actor{a}_i)$. One important property of an SDFG is \textit{throughput}, which is defined as the inverse of its long-term period. A period is the average time needed for one iteration of the SDFG. An iteration is defined as the minimum non-zero execution such that the original state of the SDFG is obtained. This is the performance parameter used in this paper. Following definitions are introduced to formulate throughput.

\begin{Definition}{Repetition Vector}
The Repetition Vector \emph{RptV} of an SDFG is defined as the vector specifying the number of times actors in the SDFG are executed in one iteration.
\end{Definition}
For the SDFG representation of a clustered SNN, 
%One of the important characteristics of SNN SDFGs is that 
all spikes generated on a channel are consumed by the destination actor. This means that all actors are fired exactly once during one iteration of the application. So, $RptV = [1 1 1 1 1 1 1]$.

\subsection{\mr{Cyclic Dependency and Deadlock Avoidance}}
\begin{figure}[h!]
	\centering
	\vspace{-5pt}
	\centerline{\includegraphics[width=0.69\columnwidth]{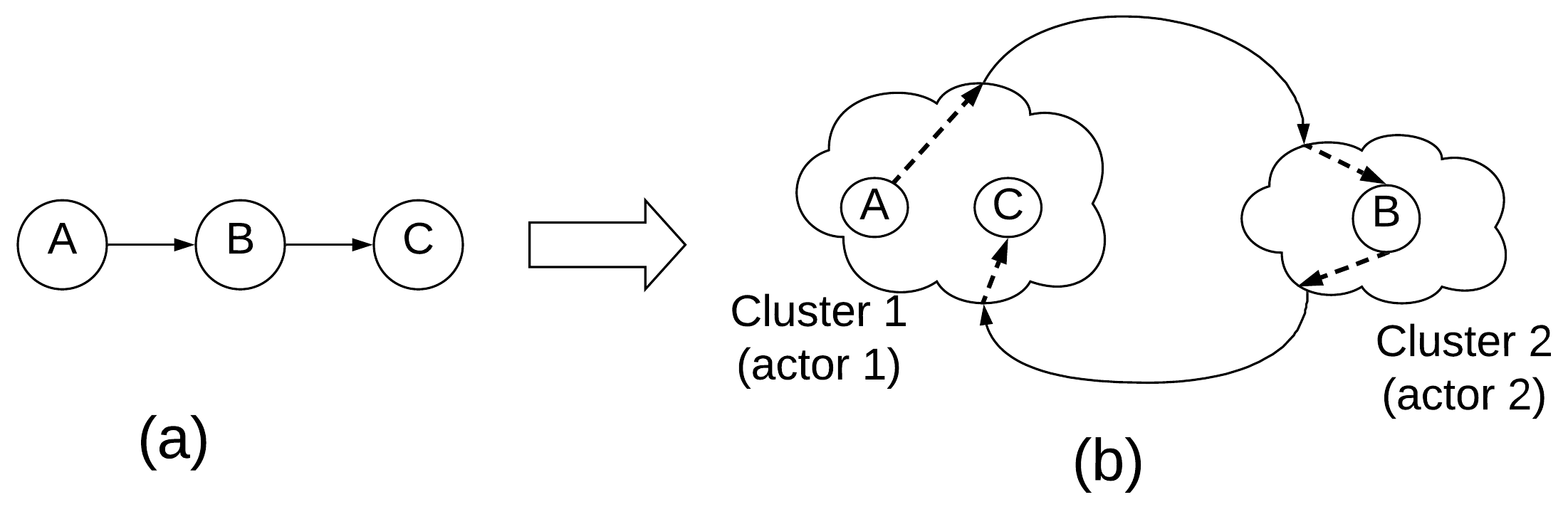}}
	%\centerline{\includegraphics[width=0.99\columnwidth]{images/LeNetMNIST_graph.png}}
	\vspace{-10pt}
	\caption{\mr{An example cycle generated during clustering of SNNs.}}
	\vspace{-10pt}
	\label{fig:cycle_example}
\end{figure}

\mr{
    The clustering approach may lead to cyclic dependency among actors. Figure \ref{fig:cycle_example}(a) illustrates a simple feedforward network of 3 neurons (A, B, \& C). Figure \ref{fig:cycle_example}(b) illustrates a scenario where neurons A and C are placed in cluster 1 (actor 1) and neuron B in cluster 2 (actor 2) during partitioning. Due to the connectivity of the neurons in Figure \ref{fig:cycle_example}(a), there is a cyclic dependency between the two actors: \underline{\texttt{actor\_1}$\rightarrow$\texttt{actor\_2}$\rightarrow$\texttt{actor\_1}}. SDF graphs allow representing such cyclic dependency among actors, justifying our choice of using them for modeling clustered SNNs.
}

\mr{
However, presence of cycles complicates the scheduling problem because cyclic dependences can lead to deadlocks. To address this, a cyclic SDF graph is decomposed into hierarchies of acyclic subgraphs. To describe this, we introduce the following definition.
%, through a process known as \emph{Cycle-Breaking} or \emph{Subindependence Partitioning}.
\begin{Definition}{Strongly Connected Subgraph}
A subgraph \ineq{Z} of a directed (cyclic or acyclic) graph 
%\ineq{\mathbf{G = (\textbf{V},\textbf{E})}} 
is called a strongly-connected subgraph, iff for every pair of vertices \ineq{a} and \ineq{b} of \ineq{Z}, there is a path from \ineq{a} to \ineq{b} and a path from \ineq{b} to \ineq{a}.
%$\tau \in \mathbb{N}\setminus\{0\}$ representing the execution time of a ($\tau(a)$).}
\end{Definition}
}

\begin{figure}[h!]
	\centering
	\vspace{-5pt}
	\centerline{\includegraphics[width=0.69\columnwidth]{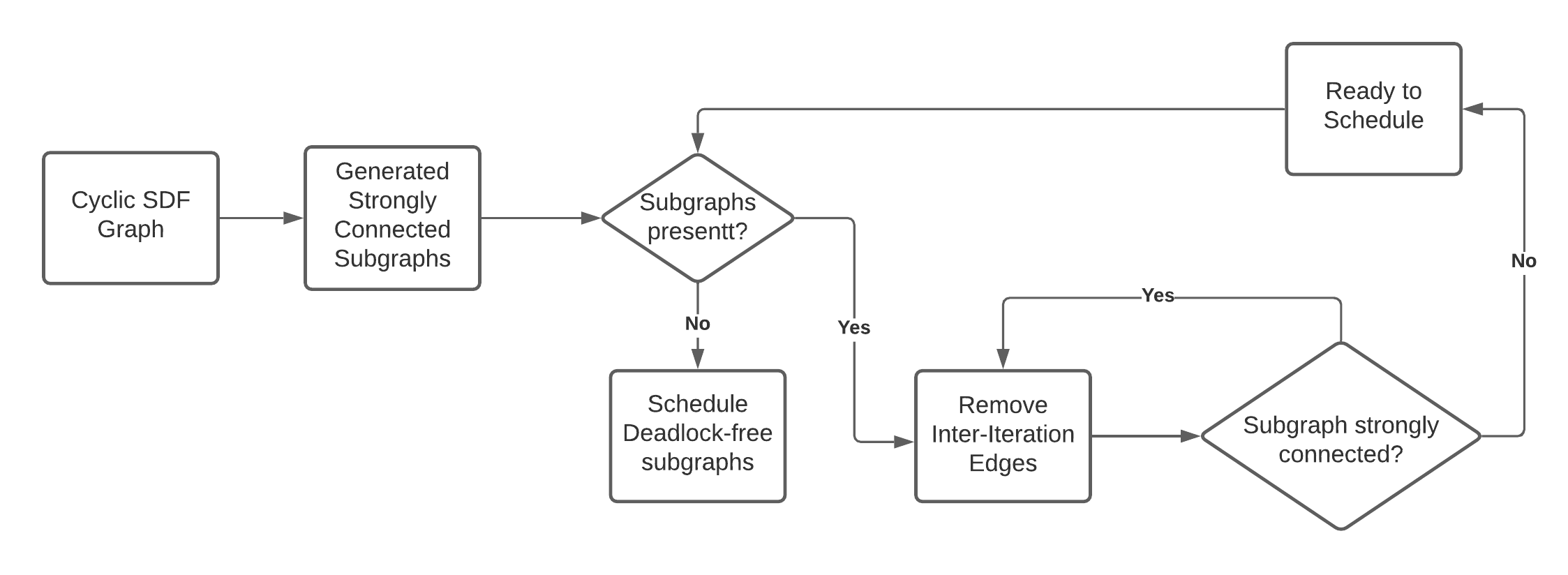}}
	%\centerline{\includegraphics[width=0.99\columnwidth]{images/LeNetMNIST_graph.png}}
	\vspace{-10pt}
	\caption{\mr{Cycle breaking for deadlock avoidance of cyclic SDF graphs~\cite{battacharyya1996loose}.}}
	\vspace{-10pt}
	\label{fig:liaf}
\end{figure}

\mr{
Figure~\ref{fig:liaf} shows the flowchart for \textit{cycle breaking}, also known as \emph{sub-independence partitioning}, which is the process of decomposition of strongly connected SDF graphs into hierarchies of acyclic graphs. This is roughly based on the Loose Interdependence Algorithms Framework (LIAF)~\cite{battacharyya1996loose}. 
A cyclic SDF graph is first decomposed into a series of strongly connected subgraphs \ineq{Z_1,Z_2,\cdots,Z_N}. For each strongly connected subgraph \ineq{Z_i}, the LIAF algorithm tries to break cycles by properly removing edges that have sufficient delays. Let \ineq{Z_i(V_i,E_i)} be the strongly-connected subgraph of the SDF Graph. An edge \ineq{e_j\in E_i} can be removed if it has enough initial tokens to satisfy the consumption requirements of its sink actor for a complete iteration of \ineq{Z_i} and scheduling \ineq{Z_i} without \ineq{e_j} does not lead to deadlock. The edge \ineq{e_j} is called \emph{inter-iteration edge}. The inter-iteration edge removal is performed iteratively until the new subgraph with the inter-iteration edges removed is no longer a strongly connected subgraph (i.e., it becomes a \emph{loosely connected subgraph}). The subgraph is pushed into a ready list for scheduling purposes. The algorithm is repeated for all the strongly-connected subgraphs. At the end, all deadlock-free subgraphs are scheduled.
}

\subsection{Performance Estimation}
We present an approach to compute the application period of an SDFG by analyzing its maximum cycle mean (MCM) and assuming infinite hardware resources.
For this, we use Max-Plus Algebra \cite{heidergott2014max,zhang2013sdc,cong2006efficient}. 
The Max-Plus semiring $\mathbb{R}_{\text{max}}$ is the set $\mathbb{R}\cup\{-\infty\}$ defined with two basic operations $\oplus \text{ and } \otimes$, which are related to linear algebra as
\begin{equation}
\label{eq:mpb}
\footnotesize a \oplus b = \max(a,b) \text{  and  } a \otimes b = a + b.
\end{equation}

\mr{
The identity element \ineq{\mymathbb{0}} for the addition \ineq{\oplus} is \ineq{-\infty} in linear algebra, i.e., \ineq{a \oplus \mymathbb{0} = a}. The identity element \ineq{\mymathbb{1}} for the multiplication \ineq{\otimes} is 0 in linear algebra, i.e., \ineq{a \otimes \mymathbb{1} = a}.
}

\begin{figure}[h!]
	\centering
	%\vspace{-5pt}
	\centerline{\includegraphics[width=0.99\columnwidth]{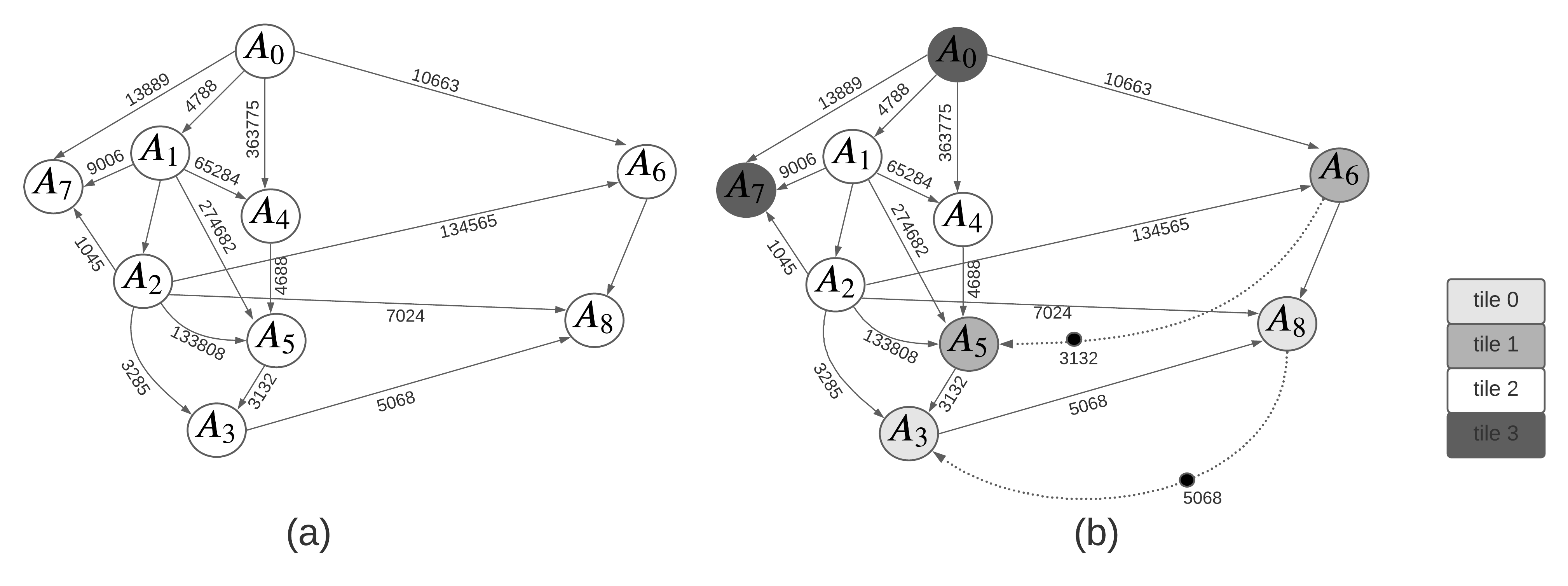}}
	%\vspace{-10pt}
	%\caption{An example of spiking neural network.}
	\caption{(a) An example of SDFG obtained from clustering of the EdgeDet application~\cite{carlsim}. (b) Mapping of the SDFG to a neuromorphic hardware with 4 tiles.}
	%\vspace{-10pt}
	\label{fig:sdfg_example}
\end{figure}

To use Max-Plus Algebra to analyze an SDFG, it is customary to express the time at which an actor fires in terms of preceding firings in linear algebra and then use standard analysis techniques for Max-Plus Algebra to estimate timing performance. We use the running example of the SDFG in Figure~\ref{fig:sdfg_example}(a), which is obtained by clustering EdgeDet~\cite{carlsim}, an application used to evaluate \tech{} (see Section~\ref{sec:evaluation}). The clustering is performed considering 1024x1024 crossbars.\footnote{We evaluate \tech{} primarily for DYNAP-SE neuromorphic hardware with $128 \times 128$ crossbars~\cite{dynapse}. Here we configure $1024 \times 1024$ crossbars to generate fewer clusters from EdgeDet for illustration purposes.} The firing end times of all 9 actors in the $k^{\text{th}}$ iteration (in linear algebra) are 

\begin{footnotesize}
\begin{align}
\label{eq:laeqn}
t_{0}(k) &\ge t_{0}(k-1) + \tau_{0} & t_{5}(k) &\ge \texttt{max}\Big[t_{2}(k),t_{1}(k),t_{4}(k)\Big] + \tau_5\\
t_{1}(k) &\ge t_{0}(k) + \tau_1 & t_{6}(k) &\ge \texttt{max}\Big[t_{2}(k),t_{0}(k)\Big] + \tau_6\nonumber\\
t_{2}(k) &\ge t_{1}(k) + \tau_2 & t_{7}(k) &\ge \texttt{max}\Big[t_{1}(k),t_{0}(k)\Big] + \tau_7\nonumber\\
t_{3}(k) &\ge \texttt{max}\Big[t_{2}(k),t_{5}(k)\Big] + \tau_3 & t_{8}(k) &\ge \texttt{max}\Big[t_{2}(k),t_{3}(k),t_{6}(k)\Big] + \tau_8\nonumber\\
%\max\Big[t_3(k-1),t_5(k-1),t_{2}(k),t_{4}(k),t_{6}(k)\Big] + \tau_{1}\nonumber\\
t_{4}(k) &\ge \texttt{max}\Big[t_{1}(k),t_{0}(k)\Big] + \tau_4 \nonumber
\end{align}
\end{footnotesize}\normalsize

Observe that the firing end time of actor \ineq{A_0} in the  $k^\text{th}$ iteration is after its firing end time in the $(k-1)^\text{th}$ iteration. Furthermore, the production and consumption rates are the same for every channel in the SDFG. Using previously introduced Max-Plus semantics, firing end times for every actor in the SDFG  can be expressed as
% \begin{equation}
% \label{eq:mpaeq}
% \footnotesize t_\mathfrak{n}(k) = \oplus t_\mathfrak{m}(k-1) \otimes \tau_\mathfrak{n}\text{, }\forall \mathfrak{m}\in Pre(\mathfrak{n})
% \end{equation}
% With a simple transformation of variables, the above sum-of-product equation can be rewritten as 
\mr{
\begin{equation}
\label{eq:mat}
\footnotesize\mathbf{t_k} = \oplus\mathbf{{T}\otimes t_{k-1}}
\end{equation}
where $\mathbf{{T}}$ is a matrix in \ineq{\mathbb{R}_{\text{max}}^{8\times 8}} that captures the actor execution times $\tau_{n}$ and \ineq{\mathbf{t_k} = \{t_0(k),t_1(k),\cdots,t_8(k)\}}. The following definitions are introduced to estimate latency.
}
\begin{Definition}{{Digraph}}
	The digraph $\Gamma(T)$ of a $n\times n$ matrix $T$ with entries defined in $\mathbb{R}_{\text{max}}$ is the tuple $\langle A,E\rangle$, where $A$ is the set of vertices, i.e., $A = \{1,2,\cdots n\}$ and $E$ is the set of connected ordered arcs between vertices i.e., $E = \{(i,j)~|~T_{i,j}\neq -\infty\}$.
\end{Definition}

\begin{figure}[h!]
	\centering
	%\vspace{-5pt}
	\centerline{\includegraphics[width=0.19\columnwidth]{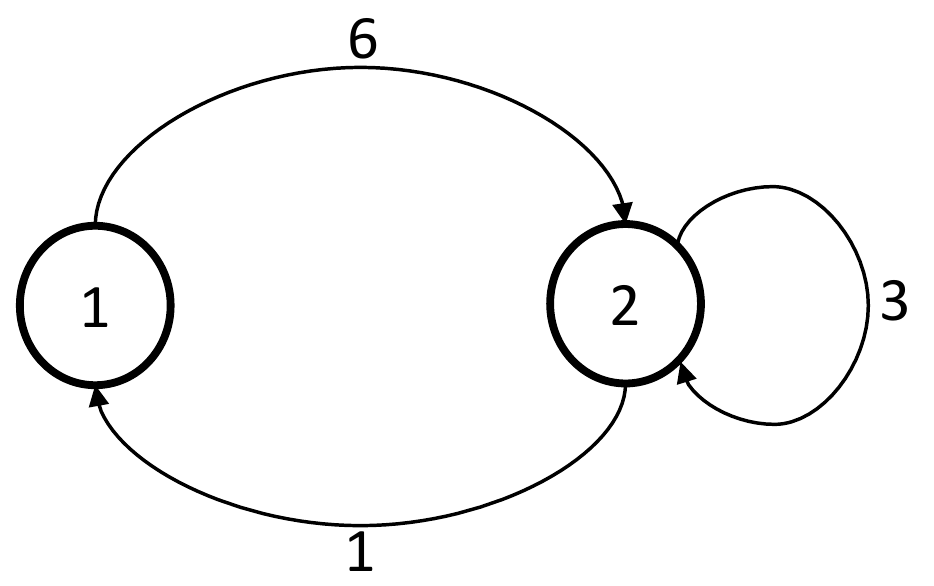}}
	%\vspace{-10pt}
	%\caption{An example of spiking neural network.}
	\caption{An example digraph.}
	%\vspace{-10pt}
	\label{fig:digraph_example}
\end{figure}

\mr{
To give an example, the matrix \ineq{T = \begin{bmatrix}
-\infty & 6\\
1 & 3
\end{bmatrix}}
corresponds to the digraph shown in Figure~\ref{fig:digraph_example}.
}

\begin{Definition}{{Walk}}
	A walk $w$ in digraph $\Gamma(T)$ is the sequence of arcs $(x_1,x_2)(x_2,x_3)\cdots(x_{k-1},x_k)$;  head of an arc in the sequence is either the start vertex of the walk or tail vertex of a preceding arc; and the tail vertex of an arc in the sequence is either the end vertex of the walk or head vertex of a succeeding arc. Weight of the walk is given by
	\begin{equation}
	\label{eq:weight}
	\footnotesize|w|_T =  T_{x_1 x_2} + \cdots T_{x_{k-1} x_k}
	\end{equation}
\end{Definition}
\begin{Definition}{{Cycle}}
	A cycle $c$ in digraph $\Gamma(T)$ is the walk $(x_1,x_2)(x_2,x_3)\cdots(x_{k-1},x_k)$, such that $x_k = x_1$.
\end{Definition}
\begin{Definition}{{Maximum Cycle Mean}}
	The maximum cycle mean, $\rho_\text{max} (T)$ is the maximum of the weight-to-length ratio of all cycles $c$ in $\Gamma(T)$ i.e.,
	\mr{
	\begin{equation}
	\label{eq:mcm}
	\footnotesize\rho_\text{max} (T) = \max\limits_{\forall c \text{ in }\Gamma(T)}\frac{|c|_T}{|c|} = \max\limits_{k > 1} \max\limits_{x_1,\cdots,x_{k-1}} \frac{T_{x_1 x_2} + \cdots T_{x_{k-1} x_k}}{k-1}
	\end{equation}
	}
	%An alternative technique to compute the maximum cycle mean is using spectral analysis \cite{cochet1998numerical}.
\end{Definition}

In this paper, \textbf{performance of an SNN is defined in terms of throughput} of the equivalent SDFG, measured as the inverse of its \textit{maximum cycle mean} (Equation~\ref{eq:mcm}), i.e.,
\mr{
\begin{equation}
    \label{eq:perf_def}
    \footnotesize \text{Performance (throughput)} = \frac{1}{\rho_\text{max} (T)}
\end{equation}
In Equation~\ref{eq:perf_def}, the performance is computed using the worst-case execution time of an actor on a crossbar. This is obtained from the propagation delay of current through the synaptic elements in the crossbar. As shown in many recent works~\cite{twisha_endurance,twisha_thermal,espine}, the current propagation delay within a crossbar depends on the specific synaptic elements that are being activated in the crossbar. This is due to the difference in the amount of parasitic components on the bitlines and wordlines of a crossbar along the different current paths. For performance guarantee purposes, we assume the worst-case propagation delay in the crossbar, and use the same to represent the execution time of actors on the crossbars of a neuromorphic hardware.
}

\mr{
The performance metric defined in Equation~\ref{eq:perf_def} provides the maximum throughput, considering only the worst-case execution time of actors. However, a neuromorphic hardware introduces constraints such as limited buffer space on the crossbars and non-zero latency on the interconnect, which can lower the throughput significantly. Therefore,
\begin{equation}
    \label{eq:throughput_snn}
    \footnotesize\text{Throughput}_{\big{|}_{SNN}} \le \text{Throughput}_{\big{|}_\text{max}} = \frac{1}{\rho_\text{max} (T)}
\end{equation}
}

\mr{
In this work, we show that performance is impacted by
\begin{enumerate}
    \item how hardware resources are allocated to actors of a clustered SNN (Section~\ref{sec:resource_allocation}), and
    \item how actors mapped to the same crossbar are time-multiplexed and scheduled (Section~\ref{sec:scheduling}).
\end{enumerate}
}

\mr{
We seek to find the lower bound on performance (\ineq{\text{Throughput}_{\big{|}_\text{bound}}}) such that
\begin{equation}
    \label{eq:lower_bound}
    \footnotesize \text{Throughput}_{\big{|}_\text{bound}} \le \text{Throughput}_{\big{|}_{SNN}} \le \text{Throughput}_{\big{|}_\text{max}}
\end{equation}
}

\mr{
By making \ineq{\text{Throughput}_{\big{|}_\text{bound}}} close to \ineq{\text{Throughput}_{\big{|}_\text{max}}}, we provide a tighter bound on performance.
}

%% file: sections/allocation.tex
The performance obtained using Equation~\ref{eq:mcm} defines the maximum throughput obtained when the clustered SNN is mapped to a hardware with infinite resources, i.e., a hardware with as many crossbars as the number of actors (clusters) in the clustered SNN graph. Additionally, each crossbar is assumed to have sufficient buffer space to send and receive spikes over the shared interconnect. However, state-of-the-art neuromorphic hardware platforms present the following three critical limitations. First, the number of crossbars in a neuromorphic hardware is limited. Therefore, the available crossbars need to be time-multiplexed amongst the clusters of an SNN. Second, the input and output buffer space on each crossbar are limited. Therefore, no more than one cluster can be executed on a crossbar concurrently. Third, the communication bandwidth of each tile is limited. Therefore, only a few spikes can be sent or received from the interconnect at once.
Formally, a neuromorphic hardware is defined as follows.

\begin{Definition}{{Neuromorphic Hardware Graph}}
	 A neuromorphic hardware graph \ineq{G_{NH} = (\textbf{T},\textbf{I})} is a directed graph consisting of a finite set \ineq{\textbf{T}} of tiles and a finite set \ineq{\textbf{I}} of interconnect links. 
\end{Definition}

Each tile consists of a crossbar to map neurons and synapses, and input and output buffers to receive and send tokens (spikes) over the interconnect, respectively. A tile \ineq{T_i} is a tuple $\langle N,inB_i,outB_i\rangle$, where \ineq{N_i} is the dimension of the crossbar on the tile, i.e., the tile \ineq{T_i} can accommodate \ineq{N_i} pre-synaptic neurons, \ineq{N_i} post-synaptic neurons, and \ineq{N_i^2} synaptic connections, \ineq{inB_i} is the input buffer size on the tile, and \ineq{outB_i} is its output buffer size. Each interconnect link is bidirectional, representing two-way communication between the source and destination tiles with a fixed bandwidth \ineq{BW}.

The mapping \ineq{\mathcal{M}: G_{CSNN} \rightarrow G_{NH}} is specified by matrix \ineq{(m_{ij}) \in \{0,1\}^{|\textbf{A}|\times|\textbf{T}|}}, 
%\rightarrow G_{nh}} is specified by a logical matrix \ineq{(m_{ij}) \in \{0,1\}^{|\mathcal{C}|\times|\mathcal{T}|}}, 
where \ineq{m_{ij}} is defined as
\begin{equation}
    \label{eq:mapping_rep}
     \footnotesize m_{ij} = \begin{cases}
    1 & \text{if actor } {A}_i \in {\textbf{A}} \text{ is mapped to tile } {T}_j\in{\textbf{T}}\\
    0 & \text{otherwise}
    \end{cases}
\end{equation}

The mapping constraint is that a cluster can be mapped to only one tile, i.e.,
\begin{equation}
    \label{eq:mapping_constraint_1}
    \footnotesize \sum_j m_{ij} = 1~~~\forall i
\end{equation}

The throughput of the clustered SNN graph \ineq{G_{CSNN}} on the neuromorphic hardware \ineq{G_{NH}} for mapping \ineq{\mathcal{M}} is computed as
\begin{equation}
    \label{eq:performance_computation}
    \footnotesize \tau_\mathcal{M} = \texttt{DFSynthesizer}(G_{CSNN},G_{NH},\mathcal{M}),
\end{equation}
where \texttt{DFSynthesizer} is the extended Max-Plus formulation of Equation~\ref{eq:mcm} incorporating platform constraints. 
The following three steps describe
%how to construct a hardware-aware SDFG (called neuromorphic SDFG) 
\texttt{DFSynthesizer}. Without loss of generality, we use Equation~\ref{eq:actor_allocation} as a running mapping example, where the 9 actors of Figure~\ref{fig:sdfg_example} are mapped to 4 tiles.
%Let the actor allocation is as follows (see Figure \ref{fig:platform_mapping}):

\begin{footnotesize}
\begin{align}
\label{eq:actor_allocation}
\texttt{\textbf{tile\_0}}: {A_3},{A_8}, &~~~~~~~~~~~~~~~~\texttt{\textbf{tile\_2}}: A_1, A_2, A_4\\
\texttt{\textbf{tile\_1}}: A_5,A_6 &~~~~~~~~~~~~~~~~\texttt{\textbf{tile\_3}}: A_0,A_7 \nonumber
%\texttt{tile\_2} &: \texttt{actor\_2}\nonumber\\
%\texttt{tile\_3} &: \texttt{actor\_1},\texttt{actor\_4}\nonumber
\end{align}
\end{footnotesize}\normalsize

The mapping corresponding to Equation~\ref{eq:actor_allocation} is therefore \ineq{\mathcal{M} = \left[
     \begin{array}{ccccccccc}
        0 & 0 & 0 & \textbf{1} & 0 & 0 & 0 & 0 & \textbf{1}\\
        0 & 0 & 0 & 0 & 0 & \textbf{1} & \textbf{1} & 0 & 0\\
        0 & \textbf{1} & \textbf{1} & 0 & \textbf{1} & 0 & 0 & 0 & 0\\
        \textbf{1} & 0 & 0 & 0 & 0 & 0 & 0 & \textbf{1} & 0\\
     \end{array}
\right]^T}.

\subsection{Step 1: Modeling Limited Buffer Sizes of Crossbars}\label{sec:step_1}
Limited input and output buffer sizes of a tile are modeled as back-edges with initial tokens indicating the buffer size available on the tile. 
This is illustrated in Figure~\ref{fig:sdfg_example}(b) with the back-edge from \ineq{A_8} to \ineq{A_3}, both of which are mapped to tile 0.
When an actor generates spikes on a channel, the available size reduces; when the receiving actor consumes the spike, the available buffer is released.
In the example, before \ineq{A_3}
%Indicating that the actor mapped to the tile can now be executed. Figure~\ref{fig:sdfg_example}(b) shows such an example of a back-edge, where the buffer size of the channel from \texttt{actor\_4} to \texttt{actor\_1} is shown as five. Before \texttt{actor\_4} 
can be executed, it has to check if enough buffer space is available. This is modeled by requiring tokens from the back-edge to be
consumed. Since it produces 5068 spikes per firing, 5068 tokens from the back-edge are consumed, indicating reservation of the buffer spaces. On the consumption side, when \ineq{A_8} is executed, it frees 5068 buffer spaces, indicated by a release of these tokens on the back-edge. 
We assume \emph{atomic} execution of actors on a crossbar, i.e., a crossbar reads input tokens and produces output tokens in the output buffer for no more than one actor at any given instance of time. To prevent other actors mapped to the same tile from firing simultaneously, 
%In this model, 
the output buffer space is claimed at the start of execution and released only at the end of firing.

\subsection{Step 2: Actor Ordering on Crossbars}\label{sec:step_2}
The number of crossbars in a neuromorphic hardware is limited. Therefore they may have to be shared between actors of an SNN. However, on a tile, only one instance of an actor can be executing at the same moment in time. We use time-division multiple-access (TDMA) to allocate time slices to actors mapped to the same tile. During its allocated time slice, an actor is executed on the crossbar of the tile and generates spikes, which are stored in the output buffer for communication on the interconnect. Next, we generate the order in which the actors bound to a tile are fired to provide performance guarantee, i.e., throughput. For this, we apply our Max-Plus Algebra formulation (Eq.~\ref{eq:mcm}) on the SDFG of Fig.~\ref{fig:sdfg_example}(b). This is our \emph{static-order schedule}, and is constructed at \textit{design time}.

\subsection{Step 3: Actor Execution on Crossbars}\label{sec:step_3}
Once the static-order schedule is constructed for all tiles of the hardware, we use a self-timed execution strategy~\cite{moreira2007self} to execute these actors at run time. Here, the exact firing times of actors are discarded, retaining only the assignment and ordering of actors on each tile as obtained from the design-time analysis (step 2). At run time, ready actors are inserted into a list and fired in the same order previously determined during design time. 

\subsection{Mapping Exploration}\label{sec:mapping_exploration}
Sections~\ref{sec:step_1} through \ref{sec:step_3} extend the Max-Plus formulation to incorporate platform constraints. 
\mr{
Using these constraints and the new formulation, one can estimate the throughput of a clustered SNN on a neuromorphic hardware for a specific actor-to-tile mapping. In the following, we explain the mapping scenario where the number of tiles in the hardware is less than the number of actors in the clustered SNN. Therefore, each tile needs to be time-multiplexed between multiple actors.
}

\begin{figure}[h!]
	\centering
	%\vspace{-5pt}
	\centerline{\includegraphics[width=0.99\columnwidth]{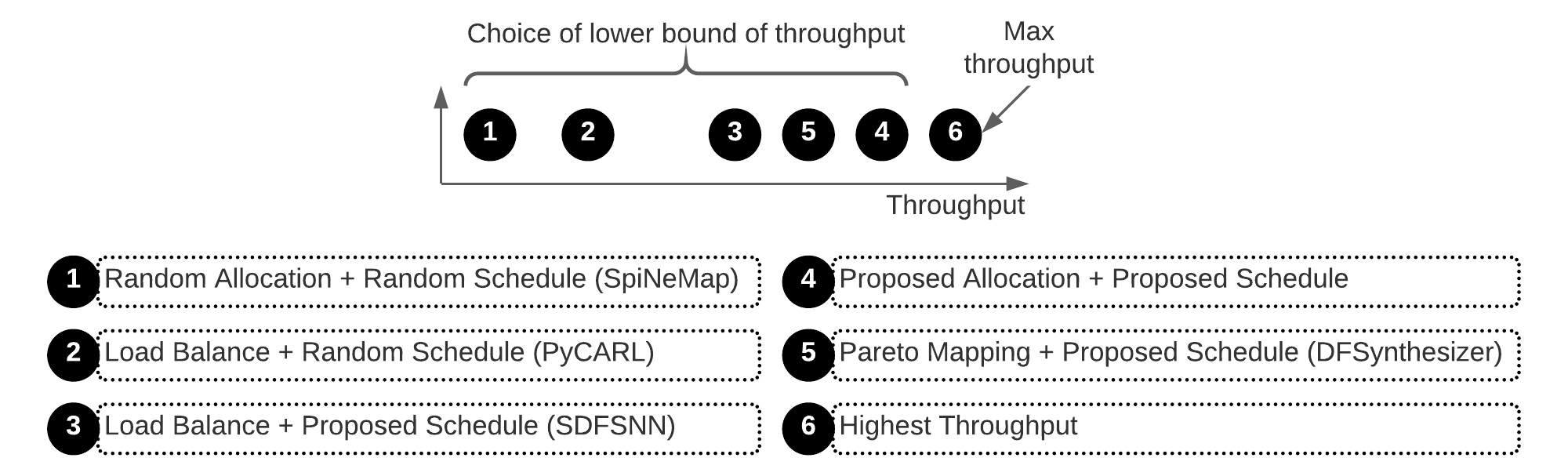}}
	%\vspace{-10pt}
	%\caption{An example of spiking neural network.}
	\caption{\mr{Different mapping explorations and choices for the lower bound of throughput (see Equation~\ref{eq:lower_bound}).}}
	%\vspace{-10pt}
	\label{fig:sota}
\end{figure}

\minor{
Figure~\ref{fig:sota} conceptually illustrates the mapping exploration using \tech{} compared to state-of-the-art solutions and the selection of lower bound on throughput.
}
\mr{
\ding{182} represents the throughput obtained using \underline{SpiNeMap}~\cite{spinemap}, which optimizes energy consumption for a hardware platform where the number of tiles is higher than the number of actors. When SpiNeMap is applied to the case where the tiles need to be time-multiplexed, it randomly distributes the actors to the tiles and schedules them arbitrarily, without considering throughput. Therefore, the throughput represented by \ding{182} (SpiNeMap) is significantly lower than the maximum throughput} \minor{(i.e., the upper bound)} \mr{represented using \ding{187}.
}
\minor{
Therefore, the throughput variation is \ineq{T_{\ding{187}} - T_{\ding{182}}}.
}

\mr{
In Figure~\ref{fig:sota}, \ding{183} represents the throughput obtained using a solution such as \underline{PyCARL}~\cite{pycarl}, which balances the load on each tile for a scenario where actors need to be time-multiplexed on the tiles. However, the actors mapped to a tile are scheduled in an arbitrary order without considering throughput. By balancing the tile load, PyCARL reduces the number of clusters mapped per tile, which improves throughput. Therefore, the throughput represented by \ding{183} is higher than \ding{182}, but lower than the maximum throughput \ding{187}. \minor{Therefore, the throughput variation is \ineq{T_{\ding{187}} - T_{\ding{183}}}.}
}

\mr{
In Figure~\ref{fig:sota}, \ding{184} represents the throughput obtained using our previous work \underline{SDFSNN}~\cite{dfsynthesizer}, which first balances the load of each tile by distributing the actors evenly, and then uses a dataflow approach to schedule the actors on each tile, improving throughput. The throughput represented by \ding{184} is therefore higher than both \ding{182} and \ding{183}, \minor{but lower than the maximum throughput \ding{187}.
Therefore, the throughput variation is \ineq{T_{\ding{187}} - T_{\ding{184}}}.}
}

\mr{
In Figure~\ref{fig:sota}, \ding{185} represents the throughput obtained using a mapping exploration framework, which explores a combination of actor-to-tile mapping and dataflow-based scheduling of actors on each tile to maximize the throughput. This throughput is higher than \ding{182}-\ding{184}, and is closer to the maximum throughput \ding{187}.
Finally, \ding{186} represents the throughput obtained using 
%\tech{}, which generates 
an actor-to-tile mapping that jointly optimizes energy and throughput, and uses dataflow-based scheduling of actors on each tile to further improve the throughput. Since this solution takes energy into consideration in the mapping step, the throughput can be somewhat lower than \ding{185} as illustrated in the figure. In Section~\ref{sec:results}, we evaluate all these approaches and show that \ding{186} is still higher than \ding{182}-\ding{184}. 
%This means that we provide a stricter performance guarantee.
}

\minor{
To conclude, the design-space exploration of \tech{} can generate mappings representing two minimum throughput solutions -- \ding{185} and \ding{186}. 
Although the maximum throughput remains the same for \tech{} and other state-of-the-art approaches, the minimum throughput of \tech{} (i.e, \ding{186}) is higher than the minimum throughput obtained using all state-of-the-art mapping solutions (i.e., \ding{182}-\ding{184}).
Therefore, the difference between maximum and minimum throughput is the least in \tech{} compared to all state-of-the-art solutions, meaning that \tech{} provides stricter performance guarantee, which is critical for real-time systems. 
}
\mr{
%By generating the lower bound on the throughput that is higher than state-of-the-art mapping solutions, we provide a stricter performance guarantee. 
We now describe \tech{}.
}

We integrate the extended Max-Plus formulation inside a design-space exploration framework to obtain cluster mappings that are Pareto optimal in terms of hardware metrics such as throughput, latency, energy, and reliability. In the following, we describe our mapping explorations considering energy and throughput. Such formulations can be trivially extended to consider other metrics.

The energy consumption \ineq{E_\mathcal{M}} of the mapping \ineq{\mathcal{M}} is measured considering the number of spikes that are generated inside each tile and the number of spikes that are routed on the interconnect~\cite{twisha_energy}. The energy parameters are reported in Table~\ref{tab:hw_parameters}. Using these parameters, the energy consumption is
\mr{
\begin{equation}
    \label{eq:energy_computation}
            \footnotesize E_\mathcal{M} = E_{spk} + E_\text{comm},
\end{equation}
% \begin{equation}
%     \label{eq:energy_computation}
%             \footnotesize E_\mathcal{M} = E_\text{event}\sum_{i=1}^{|T|}S(T_i) + E_\text{routing} \sum_{\substack{T_i,T_j\in T\\
%     i\ne j}} S(I_{i,j}),
% \end{equation}
where \ineq{E_{spk}} is the energy consumed in generating the spikes and propagating the spike current via the synapses, and \ineq{E_{comm}} is the energy consumed in communicating spikes via the shared interconnect.
}
where \ineq{S(T_i)} is the number of spikes generated inside tile \ineq{T_i\in T} and \ineq{S(I_{i,j})} is the number of spikes communicated on the link \ineq{I_{i,j}} between tiles \ineq{T_i} and \ineq{T_j} in the hardware.

Our objective is to maximize throughput of a given machine-learning model on hardware (Eq.~\ref{eq:mcm}) and minimize the hardware energy consumption (Eq. \ref{eq:energy_computation}). We formulate a joint metric \ineq{\lambda = E/\tau}, and minimize it during our mapping explorations. 
To this end, we propose an iterative approach, which explores different mapping alternatives, satisfying the cluster mapping constraint (Eq.~\ref{eq:mapping_constraint_1}). For each mapping alternative, we evaluate throughput and energy consumption. Finally, Pareto-optimal mappings are retained and returned. 

Algorithm~\ref{alg:mapping} provides the pseudo-code of our proposed mapping exploration. We start by randomly distributing clusters to the tiles (line 3). We evaluate throughput and energy consumption of this mapping and compute the joint metric \ineq{\lambda} (lines 4--5). For each cluster, we do the following. We move the cluster from its current tile to every other tile and recalculate \ineq{\lambda} (lines 6--10). If \ineq{\lambda} reduces, the new mapping is retained (lines 11--13), and the algorithm proceeds to analyze the next cluster. In this way, a local minimum is reached, starting from the initial random allocation of clusters. We re-execute the algorithm \ineq{\eta} times, starting with a different random allocation of the clusters each time. In this way, many mappings are explored. Finally, mappings that are Pareto-optimal in terms of throughput and energy consumption are retained. %The mapping that gives the highest performance and the least energy consumption is returned.

\begin{algorithm}[h]
	\scriptsize{
 		\KwIn{\ineq{G_{cl}= (C,A),G_{nh}= (T,I)}}
 		\KwOut{\ineq{\mathcal{M}_\text{max}}}
 		$\mathbb{M} = \{\}$\tcc*[r]{This set holds all the mappings}
 		\For(\tcc*[f]{Run for $\eta$ times}){$r=0;r<\eta;r\texttt{++}$}{
 		    Allocate clusters randomly to tiles. Call this mapping $\mathcal{M}$\;
 		    Calculate $\tau_\mathcal{M}$ using (\ref{eq:mcm}) and energy consumption $E_\mathcal{M}$ using (\ref{eq:energy_computation})\;
 		    Calculate the joint metric $\lambda = \tau_\mathcal{M}\cdot E_\mathcal{M}$\;
 		    \For(\tcc*[f]{For each cluster in the graph $G_{cl}$}){$C_i\in C$}{
 		        $T_{C_i} = \texttt{GetTileOfCluster}(\mathcal{M,C_i})$\tcc*[r]{Get the tile to which the cluster $C_i$ is mapped in the mapping $\mathcal{M}$}
 		        \For(\tcc*[f]{Move the cluster to every other tile }){$T_j\in T\setminus T_{C_i}$}{
 		            $\mathcal{M}_j = \texttt{MoveClusterToTile}(\mathcal{M},C_i,T_j)$ \tcc*[r]{Update the mapping to reflect the movement of cluster $C_i$ to tile $T_j$}
 		            Calculate $\tau_{\mathcal{M}_j}, E_{\mathcal{M}_j}, \text{ and } \lambda_j$\;
 		            \uIf(\tcc*[f]{If the joint metric improves}){$\lambda_j < \lambda$}{
 		                $\mathcal{M} = \mathcal{M}_j$\tcc*[r]{Retain the new mapping}
 		            }
 		        }
 		    }
 		    $\mathbb{M}.\texttt{insert}(\mathcal{M})$
 		}
 		$\mathbb{M}_{PO} = \texttt{ParetoOptimization}(\mathbb{M})$\tcc*[r]{Retain only the Pareto-Optimal Mappings}
 		Return $\mathcal{M}_\text{max}$, the mapping with minimum execution time.
 	}
	\caption{Mapping of the clustered graph $G_{cl}$.}
	\label{alg:mapping}
\end{algorithm}

The complexity of this algorithm is as follows. The unit function \ineq{\texttt{GetTileofCluster}} is essentially an \texttt{argmax} function %computed as 
%\ineq{\texttt{argmax} \mathcal{M}(i,:])} 
with a complexity of \ineq{O(|T|)}. The unit function \texttt{MoveClusterToTile} is an update of matrix and can be performed in \ineq{O(1)}. Therefore, the complexity of the algorithm is \ineq{\eta\times|C|\times|T|}. Here, \ineq{\eta} is a user-defined parameter and controls the compilation time with a trade-off on the solution quality, i.e., execution time and energy consumption of the application on hardware.

%% file: sections/scheduling.tex
Self-timed execution is widely used to schedule SDFGs~\cite{ghamarian2006throughput}. Static schedules are constructed using worst-case actor execution times determined during design time. Actor ordering on each tile is retained while discarding the timing information. At run time, actors are fired while maintaining the same order as determined during design time. In this regard, the following lemmas are stated~\cite{ghamarian2006throughput,das2012energy,das2014communication}.

\begin{Lemma}{}
For a consistent and strongly connected SDFG, the self-timed execution consists of a transient phase followed by a periodic phase.
\end{Lemma}
\begin{Lemma}{}
For a consistent and strongly connected SDFG, the throughput of an actor is given by the average firing of the actor per unit time in the periodic phase of the self-timed execution.
\end{Lemma}

\begin{figure}[h]
\centering
\includegraphics[width=0.4\columnwidth]{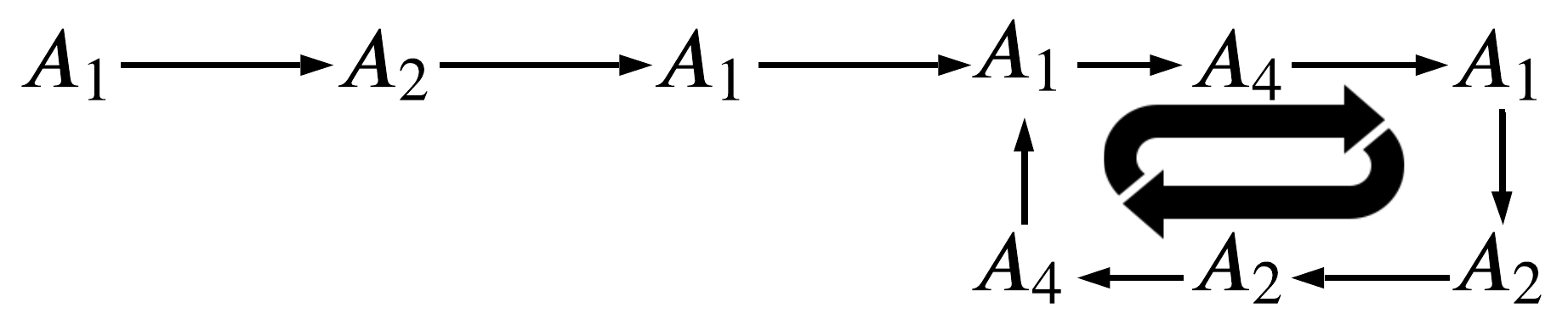}
\caption{Self-timed execution consisting of transient phase followed by periodic phase}
\label{fig:ste}
\end{figure}

Figure~\ref{fig:ste} shows an example self-timed
execution of 3 actors – \ineq{A_1}, \ineq{A_2} and \ineq{A_4} of Figure~\ref{fig:sdfg_example}(b) on tile~2.

A modern neuromorphic hardware is expected to execute many SNN applications simultaneously. When a new application is to be admitted to a hardware, which is currently running other applications, the incoming application needs to be compiled and mapped to the hardware within a short time window, based on resources currently available on the hardware. Furthermore, when an existing application finishes execution, its hardware resources are freed, meaning that such resources can now be allocated to other running applications to improve their performance. For such dynamic scenarios, SDFG schedules must be constructed for every allocation scenario. If the run-time schedule is different from that used for analysis at design time, the throughput obtained will be significantly different than what is guaranteed at design time. There are therefore two approaches to generating run-time schedules.
\begin{itemize}
\item Store the actor mapping and scheduling for all resource-allocation scenarios and for all applications from design time (\emph{storage-based} solution).
\item Construct the schedule at run time based on the mappings stored from the design-time (\emph{construction-based} solution)
%\item self-timed execution schedule (proposed)
\end{itemize}

The former is associated with high storage overhead and the latter with longer execution time. Both storage and schedule construction time are crucial for machine-learning systems deployed in resource- and power-constrained environments. Therefore, we propose a modification of the self-timed execution scheduling as follows.
First, we construct the static-order schedule for all actors of an SNN on a single tile at design time. This is achieved using the Max-Plus Algebra formulation of Equation~\ref{eq:mcm}. Next, we discard the exact timing information, retaining only the actor firing orders for run-time use.
At run time, we first construct the cluster mapping to tiles (Section~\ref{sec:mapping_exploration}), considering the available tiles. Next, we use the single-tile static-order schedule to \textbf{derive} the actor schedules on each tile, without having to construct them from scratch. %fire actors when they are ready.

\begin{figure}[h!]
	\centering
	\vspace{-5pt}
	\centerline{\includegraphics[width=0.85\columnwidth]{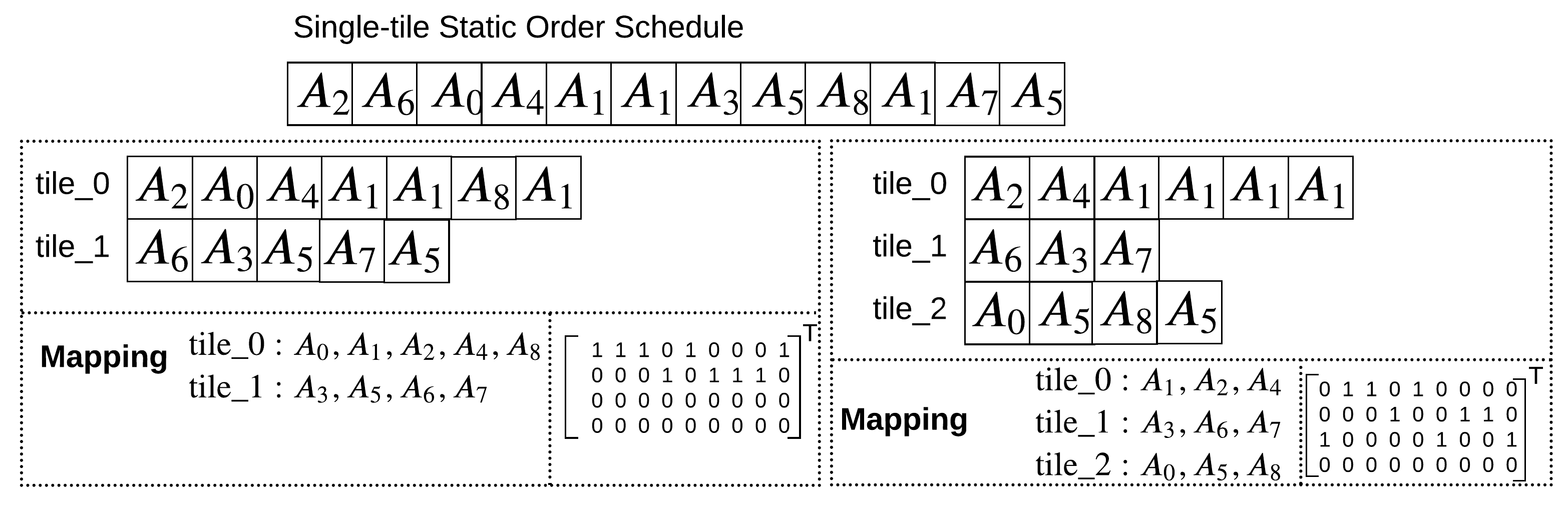}}
	\vspace{-10pt}
	\caption{Schedules constructed from the same single-tile static order schedule using 2 and 3 tiles, respectively.}
	\vspace{-10pt}
	\label{fig:schedule}
\end{figure}

Figure~\ref{fig:schedule} illustrates the construction of per-tile schedules for an SNN application with 9 actors, and with two different mappings of actors to tiles from the same single-tile static order schedule. We illustrate two scenarios in this example. In the first scenario (left), the application uses two tiles of the hardware. In the second scenario (right), the application uses three tiles of the hardware. In both scenarios, actor orders on each tile are the same as those on the single tile. Since tile schedules are not constructed from scratch, the schedule construction time is much lower.

However, performance obtained using this single-tile schedule can be lower than the maximum performance of a multi-tile schedule constructed independently. As long as this performance deviation is bounded, the actor schedule for any tile can be easily derived from the binding of actors to this tile and a given single-tile static-order schedule. See Section~\ref{sec:results} for performance evaluation.

%% file: sections/evaluation.tex
\mr{
We conduct all simulations on a Lambda workstation, which has AMD Threadripper 3960X with 24 cores, 128 MB cache, 128 GB RAM, and 2 RTX3090 GPUs. Keras~\cite{keras} and CARLsim~\cite{carlsim} use the two GPUs to accelerate model training and SNN function simulation, respectively.
}
%a system with 8 CPUs, 32GB RAM, and NVIDIA Tesla GPU, running Ubuntu 18.04.

\begin{figure}[h!]
	\centering
	\vspace{-5pt}
	\centerline{\includegraphics[width=0.85\columnwidth]{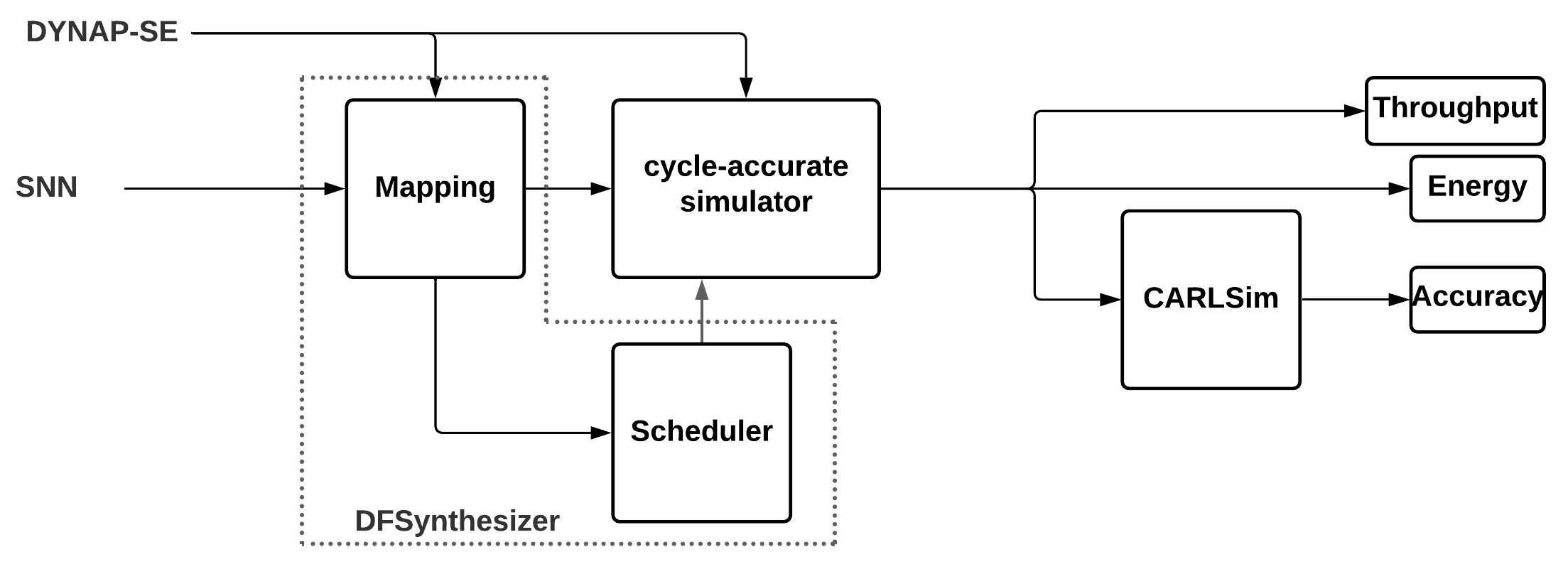}}
	\vspace{-10pt}
	\caption{\minor{Our evaluation setup based on NeuroXplorer~\cite{neuroxplorer}.}}
	\vspace{-10pt}
	\label{fig:validation_framework}
\end{figure}

\minor{
Figure~\ref{fig:validation_framework} illustrates our evaluation setup using the cycle-accurate NeuroXplorer~\cite{neuroxplorer} framework. 
This framework is validated extensively against the DYNAP-SE neuromorphic hardware~\cite{jolpe18,spinemap,psopart,HeartEstmNN,pycarl}, and can model the architecture of other neuromorphic hardware platforms such as Loihi~\cite{loihi} and TrueNorth~\cite{truenorth}. NeuroXplorer
can simulate multi-compartment neuron models and 9-parameter Izhikevich and leaky integrate-and-fire (LIF) spiking neuron models. Additionally, NeuroXplorer can model Non-Volatile Memory (NVM) synapses such as Phase Change Memory (PCM) and Oxide-based Resistive Random Access Memory (OxRRAM). NeuroXplorer also models the spike delay on the shared interconnect as well as the delay in propagating spikes through the synapses of a crossbar~\cite{neuroxplorer}.
The mapping and scheduling results obtained using \tech{} are used in NeuroXplorer to estimate energy, accuracy, and throughput. 
}

\subsection{Evaluated Applications}
We evaluate 10 machine learning programs which are representative of three most commonly-used neural network classes: convolutional neural network (CNN), multi-layer perceptron (MLP), and recurrent neural network (RNN). These applications are 
1) LeNet based handwritten digit recognition with \ineq{28 \times 28} images of handwritten digits from the MNIST dataset;
2) AlexNet for ImageNet classification;
3) VGG16, also for ImageNet classification;
4) ECG-based heart-beat classification (HeartClass)~\cite{jolpe18,das2018heartbeat} using electrocardiogram (ECG) data;
5) {image smoothing} (ImgSmooth)~\cite{carlsim} on $64 \times 64$ images; 
6) {edge detection} (EdgeDet)~\cite{carlsim} on $64 \times 64$ images using difference-of-Gaussian; 
7) {multi-layer perceptron (MLP)-based handwritten digit recognition} (DigitRecogMLP)~\cite{Diehl2015} using the MNIST database;
8) {heart-rate estimation} (HeartEstm)~\cite{HeartEstmNN} using ECG data;
9) RNN-based predictive visual pursuit (VisualPursuit)~\cite{Kashyap2018}; and
10) recurrent digit recognition (DigitRecogSTDP)~\cite{Diehl2015}.
\minor{
To demonstrate the potential of \tech{}, we consider a real-time neuromorphic system, where these machine learning programs are executed continuously in a streaming fashion. Therefore, by optimizing throughput, \tech{} improves real-time performance.
}

\mr{
Table~\ref{tab:apps} summarizes the topology, the number of neurons and synapses of these applications, and their baseline accuracy on the DYNAP-SE neuromorphic hardware using the SpiNeMap~\cite{spinemap} mapping framework.
As reported in many recent works~\cite{psopart,spinemap,pycarl}, spike latency on the shared interconnect of a neuromorphic hardware can lead to inter-spike interval (ISI) distortion and spike disorder. Since the performance of an SNN is a function of ISI, such non-idealities can lead to accuracy loss. Therefore, the accuracy of the three CNN architectures -- LeNet, AlexNet, and VGG16 in Table~\ref{tab:apps} is somewhat lower than that reported via functional simulation in Table~\ref{tab:conversion_accuracy}.
}

% \fixa{
% To demonstrate the potential of \tech{}, we enable STDP-based weight updates~\cite{kheradpisheh2018stdp} in each of these applications.
% %\footnote{Spike-Timing Dependent Plasticity (STDP)~\cite{dan2004spike} is a learning mechanism in SNNs, where the synaptic weight between a pre- and a post-synaptic neuron is updated based on the timing of pre-synaptic inputs relative to the post-synaptic spike.} 
% But our approach is not limited to STDP. 
% }
\vspace{-10pt}
\begin{table}[h!]
	\renewcommand{\arraystretch}{0.8}
	\setlength{\tabcolsep}{2pt}
	\caption{Applications used to evaluate \tech{}.}
	\label{tab:apps}
	\vspace{-10pt}
	\centering
	\begin{threeparttable}
	{\fontsize{6}{10}\selectfont
	    %\vspace{-10pt}
		\begin{tabular}{ccc|ccl|c}
			\hline
			\textbf{Class} & \textbf{Applications} &
			\textbf{Dataset} &
			\textbf{Synapses} & \textbf{Neurons} & \textbf{Topology} & \textbf{\minor{Top-1 Accuracy (\%)}}\\
			\hline
			\multirow{4}{*}{CNN} & LeNet & MNIST & 282,936 & 20,602 & CNN & 85.1\%\\
			& AlexNet & ImageNet & 38,730,222 & 230,443 & CNN & 69.8\%\\
			& VGG16 & ImageNet & 99,080,704 & 554,059 & CNN & 90.7 \%\\
			& HeartClass~\cite{jolpe18} & Physionet & 1,049,249 & 153,730 & CNN & 63.7\%\\
			\hline
			\multirow{3}{*}{MLP} & ImgSmooth \cite{carlsim} & CARLsim & 9,025 & 4,096 & FeedForward (4096, 1024) & 100\%\\
			& EdgeDet \cite{carlsim} & CARLsim & 114,057 &  6,120 & FeedForward (4096, 1024, 1024, 1024) & 100\%\\
			& DigitRecogMLP & MNIST & 79,400 & 884 & FeedForward (784, 100, 10) & 91.6\%\\
			\hline
 			\multirow{3}{*}{RNN} & HeartEstm \cite{HeartEstmNN} & Physionet & 66,406 & 166 & Recurrent Reservoir & 100\%\\
 			& VisualPursuit \cite{Kashyap2018} & \cite{Kashyap2018} & 163,880 & 205 & Recurrent Reservoir & 47.3\%\\
 			& DigitRecogSTDP \cite{Diehl2015} & MNIST & 11,442 & 567 & Recurrent Reservoir & 83.6\%\\
			%Input(32x32x3) - [Conv, Pool]*6 - [Conv, Pool]*6 - FC*84 - FC*10
			\hline
	\end{tabular}}
	%}
% 	\begin{tablenotes}\scriptsize
%         \item[a.] Input(24x24) - [Conv, Pool]*16 - FC*150 - FC*10
%         \item[b.] Input(32x32) - [Conv, Pool]*6 - [Conv, Pool]*16 - Conv*120 - FC*84 - FC*10
%         \item[c.] Input(32x32x3) - [Conv, Pool]*6 - [Conv, Pool]*6 - FC*84 - FC*10
%         \item[d.] Input(82x82) - [Conv, Pool]*16 - [Conv, Pool]*16 - FC*256 - FC*6
%     \end{tablenotes}
	\end{threeparttable}
	%\vspace{12pt}
	\vspace{-10pt}
\end{table}

\subsection{Hardware Parameters}
\minor{We model the DYNAP-SE neuromorphic hardware~\cite{dynapse} with 1024 tiles organized in a $32\times 32$ mesh.} Each tile has one $128 \times 128$ crossbar. To test the scalability of \tech{}, we also evaluate other crossbar configurations, e.g., $256 \times 256$, $512 \times 512$, and $1024 \times 1024$. Table~\ref{tab:hw_parameters} reports the relevant hardware parameters.

\begin{table}[h!]
    \caption{Major simulation parameters extracted from \cite{dynapse}.}
	\label{tab:hw_parameters}
	\vspace{-10pt}
	\centering
	{\fontsize{6}{10}\selectfont
		\begin{tabular}{lp{5cm}}
			\hline
			Neuron technology & 28nm FD-SOI\\
			\hline
			Synapse technology & \mr{HfO${}_2$ -based OxRAM}\\
			\hline
			Supply voltage & 1.0V\\
			\hline
			Energy per spike & 50pJ at 30Hz spike frequency\\
			\hline
			Energy per routing & 147pJ\\
			\hline
			Switch bandwidth & 1.8G. Events/s\\
			\hline
% 			NVM related & 1T-1R\\
% 			& PCM cell SET: 24 cycles\\
% 			& PCM cell RESET: 18 cycles\\
% 			& Program \& Verify: 35 cycles\\
% 			\hline
	\end{tabular}}
\end{table}

\minor{
The additional overhead in time multiplexing the tiles among multiple crossbars is incorporated in computing the throughput using NeuroXplorer. Specifically, once the cluster mapping to tiles are generated using \tech{}, the synaptic weights of all clusters mapped to a tile are pre-loaded into the tile's local memory (see our system architecture in Figure~\ref{fig:system_architecture}). In this way, \tech{} reduces the overhead of transferring synaptic weights at run-time from the shared main memory. Additionally, since the loading of clusters (context switching) in crossbars happen concurrently from their respective private memory, the time-multiplexing overhead is minimal.    
}

\subsection{Evaluated Metrics}
We evaluate the following performance metrics.

\begin{itemize}
    \item \textbf{Performance.} This is the throughput of each application on the hardware.
    \item \textbf{Resource Utilization.} This is the neuron, synapse, buffer, connection, and input and output bandwidth utilization on the hardware for each application.
    \item \textbf{Energy Consumption.} This is the energy consumed on the hardware for each application. \mr{This is the total energy consumed to generate spikes on each tile and communicate spike between tiles via the shared interconnect.}
    \item \textbf{Cluster Connection.} This is the average degree of the SDFG as percentage of the total number of nodes, obtained using the clustering technique for each application.
    \item \textbf{Spike Communication.} This is the total number of spikes communicated on the shared interconnect of the neuromorphic hardware.
    \item \textbf{Synthesis Time.} This is the time to compile and map each application on the hardware.
\end{itemize}

\subsection{Evaluated Approaches}
We evaluate the following approaches.
\begin{itemize}
    \item \textbf{\sm{~\cite{spinemap}}.} This approach first partitions an SNN into clusters of neurons and synapses by incorporating its workload. The objective is to minimize inter-cluster communication. Clusters are then mapped to tiles while minimizing spike communication on the shared interconnect and reducing energy consumption. When mapping SNNs to neuromorphic hardware with fewer tiles than the number of actors, 1) \sm{} allocates actors to tiles randomly and 2) \sm{} schedules the actors on each tile arbitrarily. Therefore, \sm{} does not consider throughput.
    \item \textbf{\pc{~\cite{pycarl}}.} This approach maps neurons and synapses to tiles of a neuromorphic hardware, balancing the number of neurons and synapses on each tile. \pc{} does not incorporate SNN workload, i.e., spikes generated by neurons in the SNN. Therefore, some tiles may end up communicating more spikes than others, i.e., those tiles become the energy bottleneck. %Additionally, spikes on the tile interconnect are not minimized. \pc{} does not guarantee throughput and neither does it address energy consumption.
    \mr{\item \textbf{SDFSNN{~\cite{dfsynthesizer}}.} This approach uses the load-balancing mapping of \pc{} to allocate actors to tiles. It uses dataflow scheduling to improve the throughput.}
    \item \textbf{\tech{}.} The proposed approach first clusters an SNN, considering its workload. The objective is to improve cluster utilization. This is done by first decomposing the SNN into homogeneous neural units with fanin-of-two. The clusters are then mapped to tiles, jointly optimizing throughput and energy consumption. \tech{} uses dataflow-based scheduling of actors to tiles to further improve the throughput.
\end{itemize}

%% file: sections/results.tex
\subsection{Throughput}\label{sec:performance_results}
\begin{figure}[h!]
	\centering
	\vspace{-5pt}
	\centerline{\includegraphics[width=0.99\columnwidth]{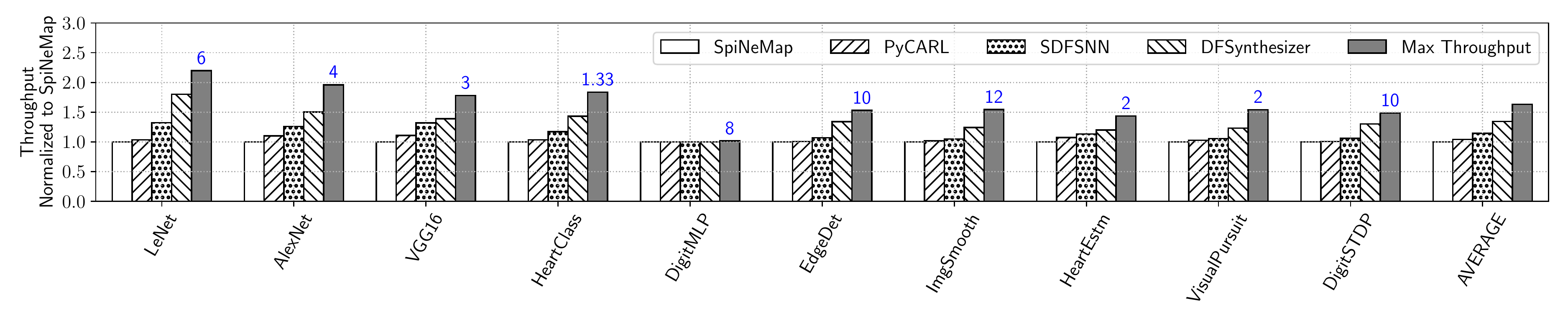}}
	\vspace{-10pt}
	\caption{Throughput on DYNAP-SE for each evaluated application normalized to \sm{}. \mr{The throughput in frames-per-second is reported for the maximum throughput approach for each application assuming unlimited hardware resources.}}
	\vspace{-10pt}
	\label{fig:throughput}
\end{figure}

Figure~\ref{fig:throughput} reports the throughput on DYNAP-SE for the evaluated approaches, for each application normalized to \sm{}. 
\mr{
For reference, we have reported the maximum throughput in frames-per-second obtained with unlimited hardware resources for each application. For image-based applications (LeNet, AlexNet, VGGNet, EdgeDet, ImgSmooth, and DigitSTDP), a frame corresponds to an individual image. For other time-series applications (HeartClass, HeartEstm, and VisualPursuit), a frame corresponds to a window of 500ms.
}
We make the following \minor{four} key observations.

\minor{
First, although the number of neurons and synapses of larger applications such as AlexNet and VGG16 is significantly higher than LeNet, the throughput of LeNet on a hardware with unlimited resources,\footnote{\minor{In the context of this work, unlimited resources refer to a neuromorphic hardware that has at least the same number of crossbars as there are clusters in the machine learning program.}} i.e., without time-multiplexing of crossbars is only 1.5x higher than AlexNet and 2x higher than VGG16. This is because with no time-multiplexing of crossbars, computations in a machine learning program take place concurrently on the crossbars, the basic philosophy of distributed computing, which is enabled using neuromorphic platforms.
Therefore, the overhead due to time-multiplexing of crossbars is no longer the throughput bottleneck. Rather, the bottleneck shifts to spike delay between the clusters. Additionally, in our framework we cluster machine learning programs to minimize inter-cluster spikes. Therefore, even though Alexnet has significantly higher number of neurons and synapses than LeNet, its number of inter-cluster spikes is not significantly higher. The throughput of AlexNet is only 33\% lower than LeNet. %thereforewhich has more inter-cluster spikes than LeNet has higher delay, resulting in a lower throughput. 
Similarly, VGG16, which has higher inter-cluster spikes than AlexNet, has 25\% lower throughput.
}

\mr{
Second, the throughput obtained using \sm{} is the least because \sm{} does not guarantee throughput during actor-to-tile mapping and actor scheduling on tiles. The throughput of \pc{} is on average 4\% higher than \sm{}. This is because \pc{} balances the load on the tiles and therefore, the average number of actors mapped to each tile is lower than \sm{}, which results in higher throughput. The throughput of SDFSNN is on average 9.7\% higher than \pc{}. This improvement is because of the use of dataflow-based scheduling, which maximizes the throughput. \tech{} improves throughput by an average of 17\% compared SDFSNN. This improvement is because unlike SDFSNN, which maps actors to tiles balancing the tile load without considering the throughput, \tech{} performs throughput- and energy-aware mapping of actors to tiles and then uses dataflow-based scheduling to further improve the throughput. We have analyzed such throughput differences in Section~\ref{sec:mapping_exploration}.}

\mr{
Third, the throughput using \tech{} is only 16\% lower on average than the maximum throughput obtained with unlimited hardware resources.
Finally, the throughput of DigitMLP is a very small application. All the techniques generate the same number of clusters for this application, resulting in similar throughput.
}

\subsection{Workload Energy}
\begin{figure}[h!]
	\centering
	\vspace{-5pt}
	\centerline{\includegraphics[width=0.99\columnwidth]{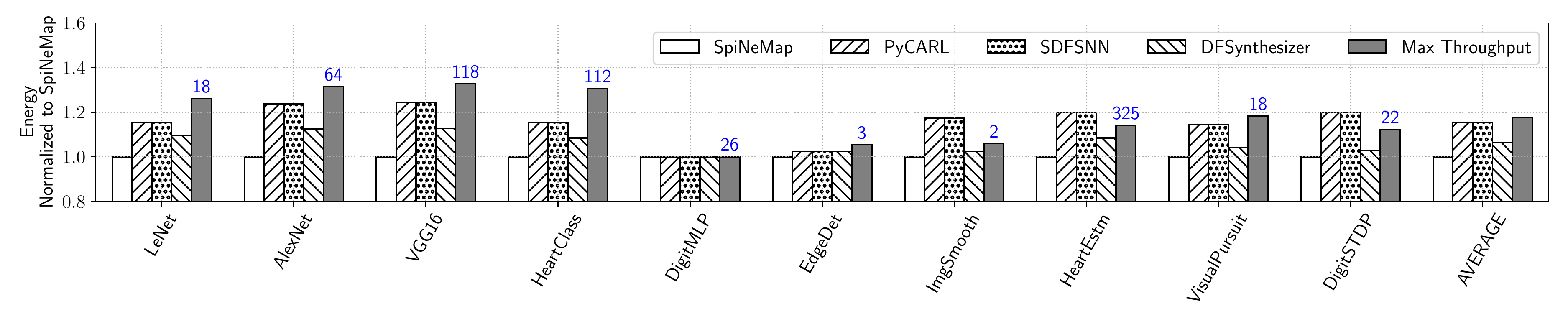}}
	\vspace{-10pt}
	\caption{Workload energy on DYNAP-SE for each evaluated application normalized to \sm{}. \mr{The workload energy in \ineq{\mu J} is reported for the maximum throughput approach for each application assuming unlimited hardware resources.}}
	\vspace{-10pt}
	\label{fig:energy}
\end{figure}

Figure~\ref{fig:energy} reports the workload energy estimated on DYNAP-SE of the evaluated approaches for each application normalized to \sm{}. 
\mr{
For reference, we have reported the workload energy in \ineq{\mu J} obtained using the maximum throughput approach, which assumes unlimited hardware resources.
}
We make the following observation.

\mr{
The energy consumption of \sm{} is the least because this approach partitions SNNs into clusters to explicitly minimize the number of inter-cluster spikes. Therefore, when the clusters are mapped to hardware, the energy consumption on the shared interconnect is reduced.\footnote{\mr{The mapping exploration only impacts the communication energy on the shared interconnect. The spike generation energy remains the same for all approaches.}} Second, the energy consumption of \pc{} is on average 15\% higher than \sm{}. This is because \pc{} balances the tile load without incorporating energy consumption. Therefore, clusters with high volume of spike communication between them may get placed on different tiles, increasing the communication energy. \sm{} places those tiles on the same tile lowering the communication energy. The energy consumption of SDFSNN is the same as \pc{} because the cluster-to-tile mapping of these two approaches is the same. SDFSNN gains over \pc{} in terms of throughput due to its dataflow-based cluster scheduling on tiles. We analyzed this in Section 8.1.  
}
\mr{
The energy consumption of \tech{} is lower than SDFSNN by an average of 8\%. This reduction is due to the cluster-to-tile mapping of \tech{}, which incorporates energy consumption.
}

\subsection{Scheduling}
\begin{figure}[h!]
	\centering
	\vspace{-5pt}
	\centerline{\includegraphics[width=0.99\columnwidth]{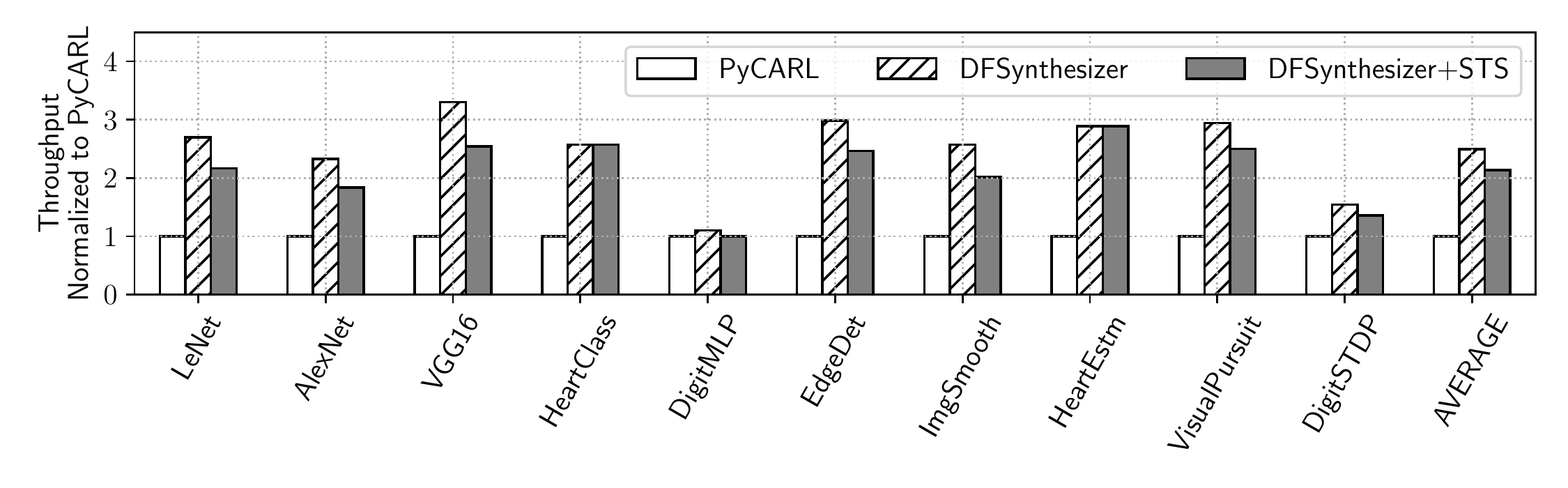}}
	\vspace{-10pt}
	\caption{Throughput normalized to \pc{}.}% for the evaluated workloads.}
	\label{fig:runtime}
	\vspace{-10pt}
\end{figure}

Figure~\ref{fig:runtime} reports throughput of each of our applications for our proposed approach normalized to \pc{}. 
We compare throughput obtained using \tech{} where schedules are independently constructed for each tile against the throughput obtained using our proposed single-tile based schedule (\tech{}+STS).
%using our approach used during compile-time, where schedule for each tile is constructed independently against throughput obtained at run-time, where schedule for each tile is constructed from a single-tile static-order schedule. 
We make the following three observations.

First, throughput obtained from a single-tile static-order schedule is on average 15\% lower than the case when schedules are constructed independently --- that is, by using \tech{}. This verifies our Lemma 2. Second, for some applications such as HeartEstm and HeratClass, throughput obtained using \tech{}+STS is exactly the same as that obtained using \tech{}. Third, throughput using \tech{}+STS is still higher than \pc{} by an average of 41\%.

\subsection{Resource Utilization}
Table~\ref{tab:resource} reports the utilization of hardware resources (tile resources, buffer size, connections, and input and output bandwidth) on the DYNAP-SE neuromorphic hardware for each application. The average utilization of hardware resources is 92.5\% for the crossbar IOs on each tile, 9.0\% for buffer space, 42.6\% for connections, and 15\% for input and output tile bandwidth.
Since we perform hardware-aware analysis, resource utilization never exceeds 100\%.

\begin{table}[h!]
	\renewcommand{\arraystretch}{1.0}
	\setlength{\tabcolsep}{3pt}
	\centering
	{\fontsize{7}{10}\selectfont
		\begin{tabular}{|l|c|c|c|c|c|}
			\hline
			\multirow{3}{*}{\textbf{Application}} & \multicolumn{5}{|c|}{\textbf{Utilization (\%)}}\\ \cline{2-6}
			& \multirow{2}{*}{\textbf{Tile}} & \multirow{2}{*}{\textbf{Buffer}} & \multirow{2}{*}{\textbf{Connections}} & 
			\multicolumn{2}{|c|}{\textbf{Bandwidth}}\\ \cline{5-6}
			&&&&\textbf{Input} & \textbf{Output}\\
			\hline
LeNet	&	100	&	87.8	&	37.5	&	20.34	&	20.34	\\
AlexNet	&	100	&	91.8	&	46.87	&	17.09	&	17.09	\\
VGG16	&	100	&	94.2	&	15.62	&	6.51	&	6.51	\\
HeartClass	&	100	&	79.1	&	25	&	9.76	&	9.76	\\
DigitMLP	&	81.25	&	9.67	&	46.87	&	22.78	&	22.78	\\
EdgeDet	&	87.5	&	11.23	&	68.75	&	22.78	&	22.78	\\
ImgSmooth	&	87.5	&	8.39	&	37.5	&	17.08	&	17.08	\\
HeartEstm	&	96.87	&	9.61	&	62.5	&	4.7	&	4.7	\\
VisualPursuit	&	90.12	&	21.2	&	25.04	&	12.11	&	16.6	\\
DigitSTDP	&	89.33	&	20.13	&	22.19	&	11.94	&	11.7	\\
			\hline
	\end{tabular}}
	%\vspace{6pt}
	\caption{Resource utilization on DYNAP-SE.}
	%\vspace{-24pt}
	\label{tab:resource}
\end{table}
%\vspace{-10pt}

These results illustrate that \tech{} can be used to design neuromorphic hardware while considering key hardware parameters such as number of tiles, but all other resources such as buffer space, connections, and input and output bandwidth.

\begin{figure}[h!]
	\centering
	\vspace{-5pt}
	\centerline{\includegraphics[width=0.99\columnwidth]{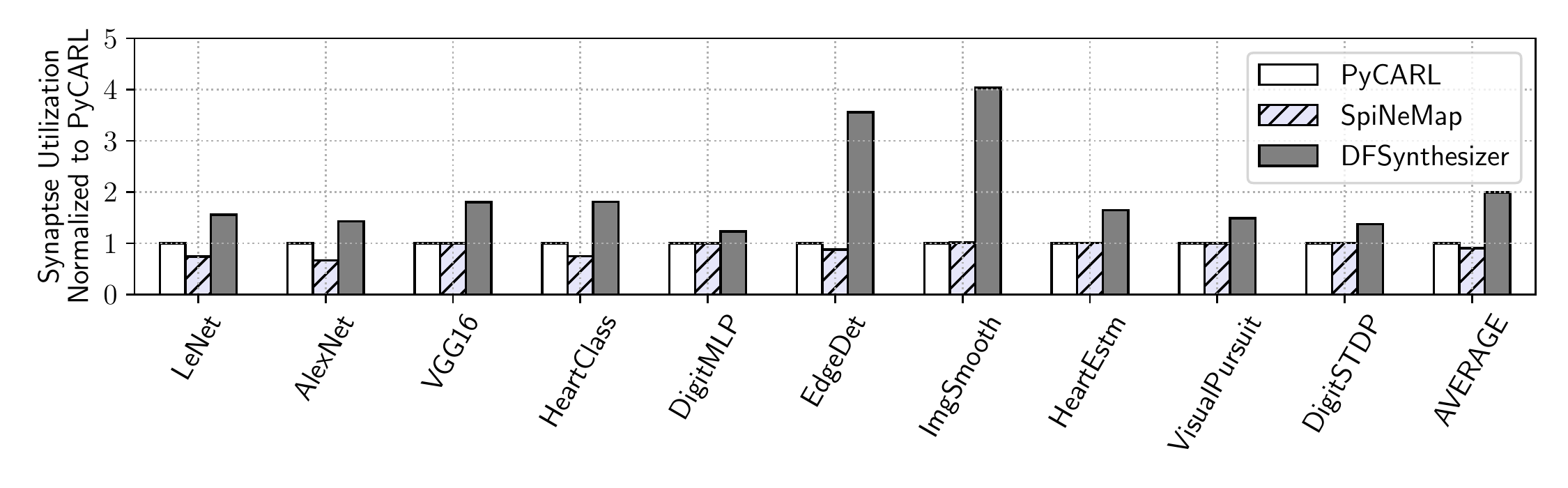}}
	\vspace{-10pt}
	\caption{Average synapse utilization on tiles for each evaluated application normalized to \pc{}.}
	\vspace{-10pt}
	\label{fig:cluster_utilization}
\end{figure}

To give more insight on the utilization within each tile, Figure~\ref{fig:cluster_utilization} reports the average synapse utilization on tiles of the evaluated approaches for each application normalized to \pc{}. We make the following two key observations.

First, the synapse utilization on tiles using \sm{} is the least of all three evaluated approaches. This is because \sm{} produces the highest number of clusters (Sec.~\ref{sec:number_of_clusters}) and therefore, the average number of synapses per cluster is the least. Subsequently, when these clusters are mapped to tiles, the average synapse utilization on tiles reduces.
Second, \tech{} generates fewer clusters than both \sm{} and \pc{} due to its dense packing of synapses using Algorithm~\ref{alg:clustering}. Therefore, the average number of synapses per cluster is higher, which increases synapse utilization on tiles when the clusters are mapped to tiles. On average, the average synapse utilization of \tech{} is 2x higher than \pc{} and 2.2x higher than \sm{}.

\subsection{Number of Clusters}\label{sec:number_of_clusters}
\begin{figure}[h!]
	\centering
	\vspace{-5pt}
	\centerline{\includegraphics[width=0.99\columnwidth]{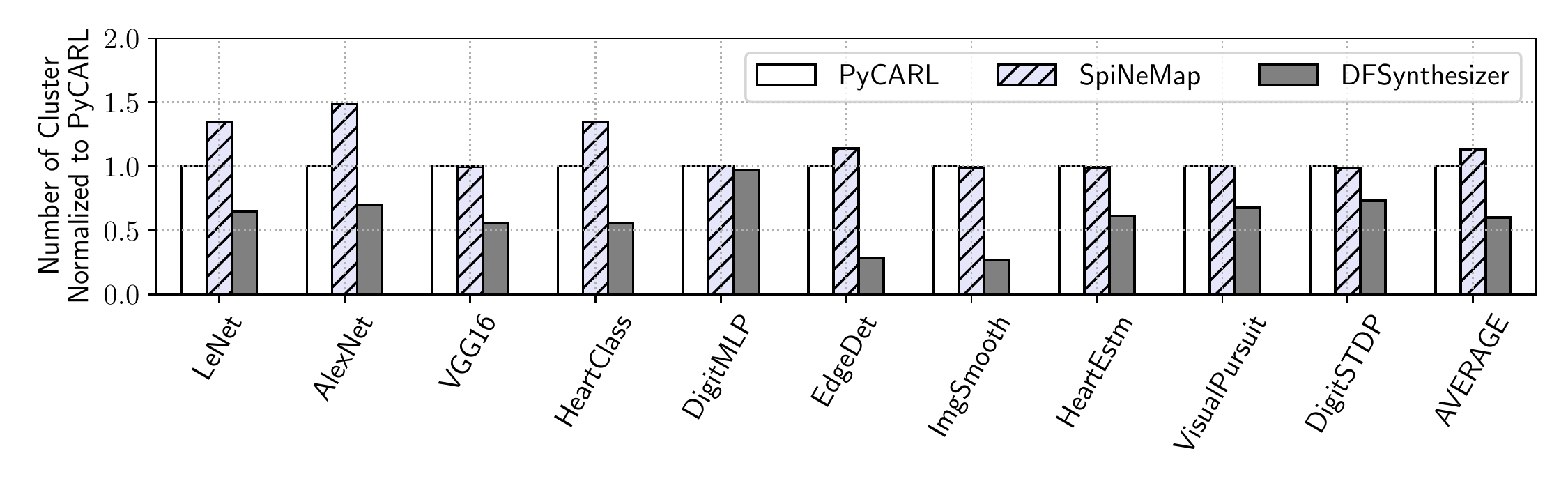}}
	\vspace{-10pt}
	\caption{Number of clusters for each evaluated application normalized to \pc{}.}
	\vspace{-10pt}
	\label{fig:total_clusters}
\end{figure}

Figure~\ref{fig:total_clusters} reports the total number of clusters of the evaluated approaches for each application normalized to \pc{}. We make the following two key observations.

First, the number of clusters of \sm{} is the highest of all three evaluated approaches. This is because \sm{} minimizes the number of inter-cluster communication during clustering of an SNN. Therefore, neurons that spike the most are placed within individual clusters along with their fanins. Since \sm{} does not consider cluster utilization, it results in creating more clusters than \pc{}. Second, \tech{} clusters an SNN to maximize the resource utilization on each tile. Therefore, the number of clusters generated by \tech{} is the lowest. Overall, the number of clusters of \tech{} is 41\% lower than \sm{} and 47\% lower than \pc{}. Lower the number of clusters, lower is the size of hardware needed to achieve highest throughput (Sec.~\ref{sec:performance_results}). Therefore, \tech{} reduces the hardware requirement for machine learning applications.

\subsection{Cluster Connections}
\begin{figure}[h!]
	\centering
	\vspace{-5pt}
	\centerline{\includegraphics[width=0.99\columnwidth]{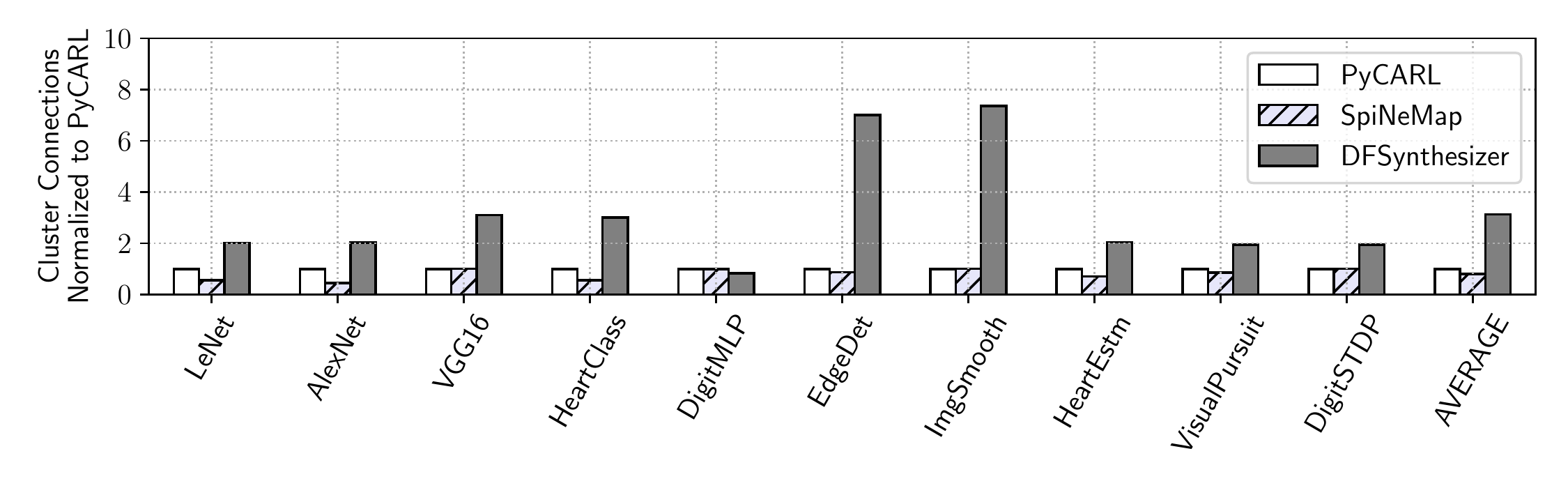}}
	\vspace{-10pt}
	\caption{Cluster connections for each evaluated application normalized to \pc{}.}
	\vspace{-10pt}
	\label{fig:cluster_connections}
\end{figure}

Figure~\ref{fig:cluster_connections} reports the cluster connections of the evaluated approaches for each application normalized to \pc{}. We make the following two key observations.

First, the number of inter-cluster connections of \sm{} is the least of all three evaluated approaches. This is because \sm{} minimizes the number of inter-cluster communication while clustering an SNN, which indirectly reduces the cluster connectivity. Second, \tech{} clusters an SNN to maximize the resource utilization on each tile. Therefore, the number of connections between the clusters is higher in \tech{} because of the higher number of post-synaptic neurons mapped to each cluster. Overall, the average cluster connections of \tech{} is 3.1x higher than \sm{} and 3.9x higher than \pc{}.

\subsection{Architecture Exploration}
\begin{figure}[h!]
	\centering
	\vspace{-5pt}
	\centerline{\includegraphics[width=0.99\columnwidth]{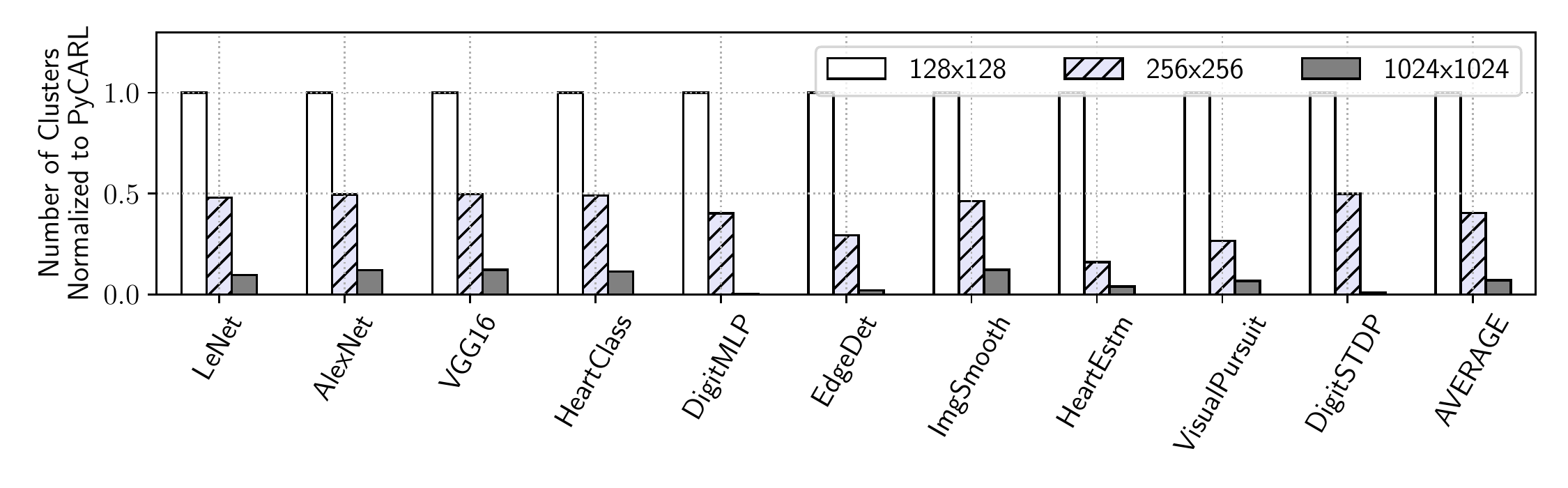}}
	\vspace{-10pt}
	\caption{Number of clusters generated using \tech{} for $128 \times 128$, $256 \times 256$, and $1024 \times 1204$ crossbars, normalized to the configuration of DYNAP-SE with $128 \times 128$ crossbars.}
	\vspace{-10pt}
	\label{fig:arch_crossbars}
\end{figure}

Figure~\ref{fig:arch_crossbars} reports the number of clusters generated using \tech{} for neuromorphic hardware with $128 \times 128$, $256 \times 256$, and $1024 \times 1024$ crossbars, normalized to a DYNAP-SE configuration with $128 \times 128$ crossbars. We observe that the number of clusters generated using \tech{} reduces by 60\% and 92\% when the size of a crossbar increases to $256 \times 256$ and $1024 \times 1024$, respectively. 

\begin{figure}[h!]
	\centering
	\vspace{-5pt}
	\centerline{\includegraphics[width=0.99\columnwidth]{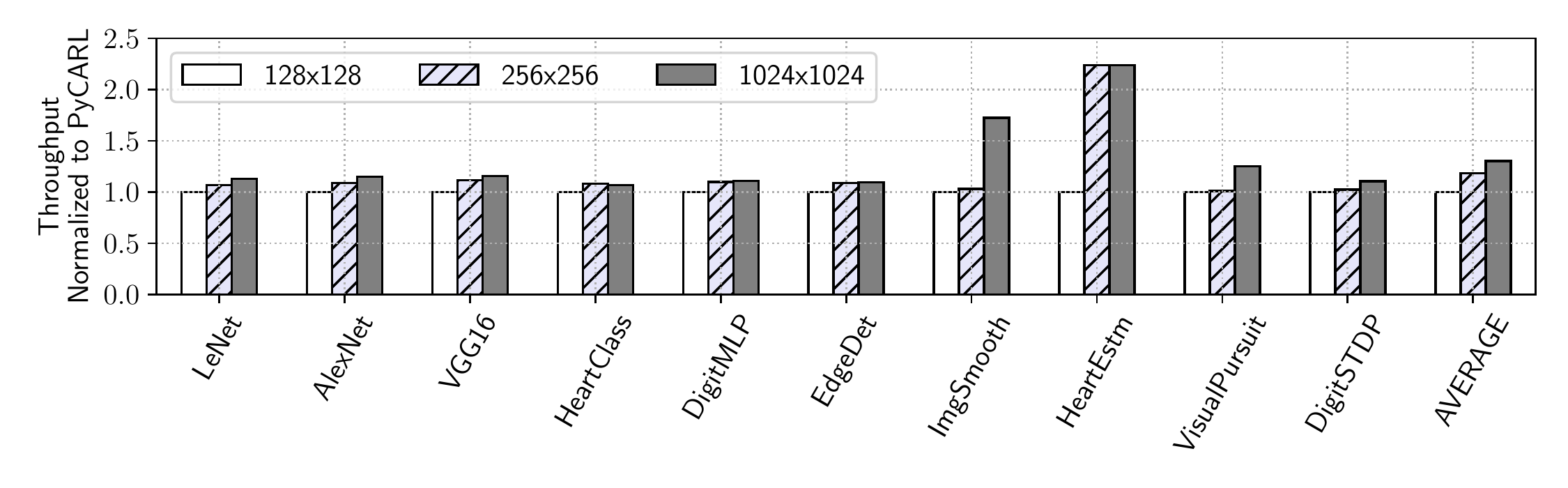}}
	\vspace{-10pt}
	\caption{Throughput achieved using \tech{} for $128 \times 128$, $256 \times 256$, and $1024 \times 1204$ crossbars, normalized to throughput on DYNAP-SE with $128 \times 128$ crossbars.}
	\vspace{-10pt}
	\label{fig:arch_throughput}
\end{figure}

Fewer number of clusters increases throughput. To illustrate this, Figure~\ref{fig:arch_throughput} reports the throughput using \tech{} for different crossbar sizes normalized to throughput on DYNAP-SE with four $128 \times 128$ crossbars. We make the following two observations.

First, throughput increases by 18\% and 30\% when using $256 \times 256$ and $1024 \times 1024$ crossbars, respectively. This improvement is because with larger size crossbars, there are fewer clusters generated by \tech{} (Fig.~\ref{fig:arch_crossbars}). Therefore, the number of clusters per tile reduces, which reduces the bottleneck of time-multiplexing clusters on tiles. This increases throughput. Second, for applications such as DigitMLP, EdgeDet, and HeartEstm, there is no throughput improvement when the crossbar size increased from $512 \times 512$ to $1024 \times 1024$. This is because for these applications, $256 \times 256$ crossbar configuration is sufficient to achieve the highest throughput. For all other applications, the throughput increases by 11\% when going from $256 \times 256$ to $1024 \times 1024$ crossbars.

\subsection{Synthesis Time}
\begin{figure}[h!]
	\centering
	\vspace{-5pt}
	\centerline{\includegraphics[width=0.99\columnwidth]{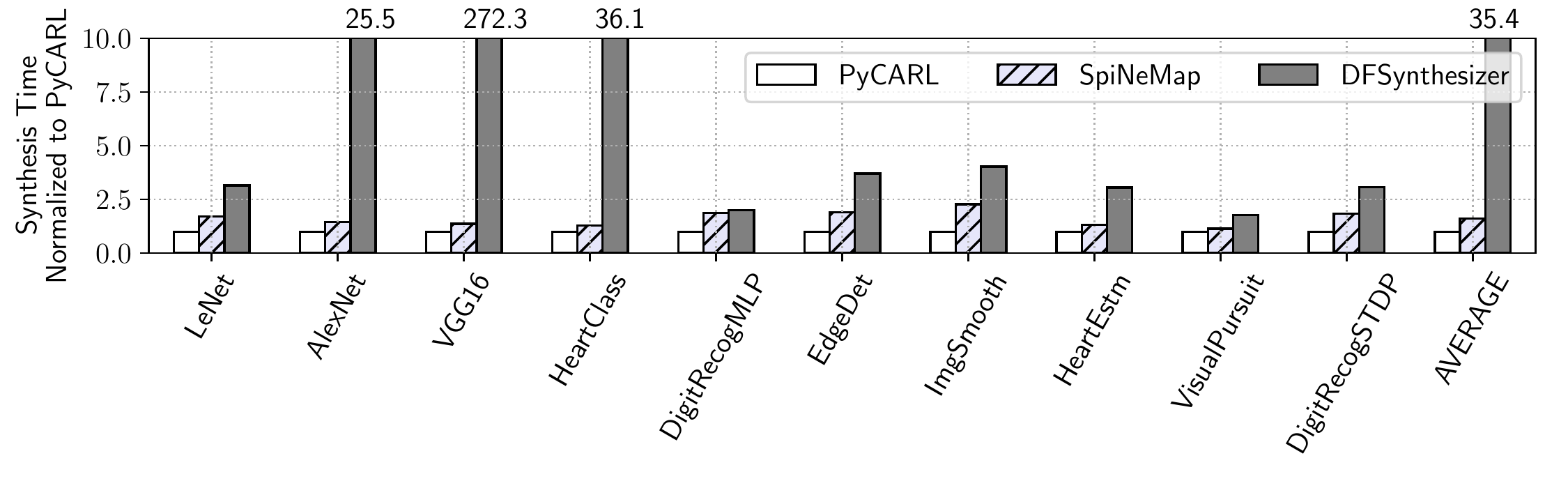}}
	\vspace{-10pt}
	\caption{Synthesis time for each application normalized to \pc{}.}
	\vspace{-10pt}
	\label{fig:compilation_time}
\end{figure}

Figure~\ref{fig:compilation_time} reports the synthesis time on DYNAP-SE for the evaluated approaches, for each application normalized to \pc{}. We make the following three key observations.

{First}, the synthesis time of \sm{} is on average 61.6\% higher than \pc{}. The higher synthesis time of \sm{} is due to the analysis it performs with the workload to obtain the minimum energy mapping. Second, the synthesis time of \tech{} is the highest. On average, the synthesis time of \tech{} is 35x higher than \pc{} and 25x higher than \sm{}. This higher synthesis time is due to 1) \tech{}'s mapping explorations using Algorithm~\ref{alg:mapping}, and 2) \tech{}'s SDFG analysis mechanism using the proposed Max Plus formulation. Third, the synthesis time of \tech{} increases with model complexity. The synthesis time of \tech{} is higher than \pc{} by 3.1x for LeNet, 25.5x for AlexNet, and 272.3x for VGG16.

% of \tech{} is better than \sm{} (throughput of \tech{} is on average {61.4\%} higher than \sm{}). This improvement is because \tech{} allocates the clusters to tiles, minimizing the performance bottleneck due to time multiplexing of fewer clusters on each tile.
% {Second}, performance of \tech{} is higher than \sm{} by 63\% for CNN-based applications, 60\% for MLP-based applications, and 61\% for RNN-based applications. 
%The performance is better for RNN-based applications because \tech{} intelligently places clusters on tiles to minimize communication on the interconnect, which reduces spike latency and hence, the execution time.
% {Third}, the number of clusters of DigitMLP generated by \tech{} is comparable to \sm{} (Sec. \ref{sec:number_of_clusters}). So, both \tech{} and \sm{} encounters similar performance bottleneck due to time multiplexing of the limited number of tiles, achieving similar performance.

\subsection{Model Quality}
\tech{} does not alter synaptic connections. Therefore, the model quality, e.g., accuracy is not impacted by the analysis technique of \tech{}. The only impact \tech{} introduces is in converting CNNs.
%to their SNN equivalent. 
The accuracy impact is reported in Table~\ref{tab:conversion_accuracy}. For all other applications, \tech{}'s accuracy is the same as the baseline accuracy reported in Table~\ref{tab:apps}.

%% file: sections/related_works.tex
\mr{
Recently, many approaches are proposed to map machine learning workloads to neuromorphic hardware.
Corelet~\cite{amir2013cognitive} is used to map SNNs to TrueNorth~\cite{truenorth}. PACMAN~\cite{galluppi2015framework} is used to map SNNs to SpiNNaker~\cite{spinnaker}. 
PyNN~\cite{pycarl} is used to map SNNs on Loihi~\cite{loihi}, BrainScaleS~\cite{schemmel2012live}, and Neurogrid~\cite{neurogrid} by balancing the load on each tile.
PyCARL~\cite{pycarl} is used to map SNNs to DYNAP-SE~\cite{dynapse}.
The primary objective of these approaches is to balance the workload on each tile by distributing the neurons and synapses evenly.
}

\mr{
Beyond load balancing, recent techniques have also explored other objectives.
PSOPART~\cite{psopart} is used to map SNNs to neuromorphic hardware, reducing the energy consumption on the shared interconnect.
\sm{}~\cite{spinemap} performs energy-aware clustering of SNNs and then maps the clusters to tiles, reducing the communication energy. DecomposeSNN~\cite{esl20} decomposes an SNN to improve the cluster utilization. There are also performance-oriented SNN mapping approaches such as~\cite{balaji2020run,dfsynthesizer,balaji2019design,adarsha_igsc}, energy-aware SNN mapping approaches such as~\cite{twisha_energy}, circuit aging-aware SNN mapping approaches such as~\cite{reneu,song2020case,balaji2019framework,vts_das,ncrtm}, endurance-aware SNN mapping approaches such as~\cite{twisha_endurance,espine,song2021improving}, and thermal-aware SNN mapping approaches such as~\cite{twisha_thermal}. These approaches are evaluated with emerging SNN based applications~\cite{moyer2020machine,jolpe18,das2018heartbeat,Diehl2015,HeartEstmNN,Kashyap2018}, which we also use to evaluate \tech{}.
}

\mr{
%\sm{} reduces the communication between tiles~\cite{spinemap}. \pc{} is proposed to perform hardware-software co-simulation of SNNs \cite{pycarl}. 
There are also other mapping approaches such as 
%reneu,twisha_thermal,twisha_endurance,esl20,balaji2020run,song2020case,balaji2019framework}, as well as approaches that use a single large crossbar to map SNNs~\cite{
~\cite{ankit2018neuromorphic,zhang2018neuromorphic,xia2019memristive,lee2019system,wijesinghe2018all,wen2015eda,ramasubramanian2014spindle}.
We compare \tech{} against \pc{} and \sm{}, and found it to perform significantly better.
}

% \subsection{Non-volatile Memory}
% Recently, mon-volatile memory such phase-change memory is used to lower the energy consumption and improve performance of von-Neumann computing systems~\cite{LeeISCA2009,palp,mneme,datacon,hebe}. NVMs can also improve the storage density and reduce energy consumption of neuromorphic computing. To this end, Ramasubramanian et al. use STT-MRAM~\cite{ramasubramanian2014spindle}, Burr et al. use PCM~\cite{Burr2017}, and Mallik et al. use OxRAM~\cite{mallik2017design} to design neuromorphic tiles. Although we demonstrate the performance and energy advantages of \tech{} using DYNAP-SE with OxRRAM synapses, our approach however, is not specific to the choice of non-volatile memory technology.

\subsection*{Similar Concept in Related Domain}
SDFGs are widely used for predictable mapping of applications to multiprocessor systems. Numerous approaches to throughput analysis of SDFGs have been previously proposed~\cite{stuijk2006exploring,stuijk2007multiprocessor,damavandpeyma2012modeling,zhu2012static,shafik2015adaptive,das2015hardware,shafik2015adaptive}. Bonfietti et al. evaluated mappings of SDFG to multiprocessor system, maximizing the throughput~\cite{bonfietti2013maximum}. Stemmer et al. propose to use probabilistic analysis to allocate and schedule SDFGs on multiprocessor systems~\cite{stemmer2020towards}. Das et al. evaluated the fault-tolerant mapping of SDFGs to multiprocessor systems~\cite{das2013communication,das2015reliability,das2014communication,das2012faultRSP,das2013aging,das2014energy,das2012energy,das2013energy,das2016adaptive}. Recently, SDFG-based analysis is also proposed for analyzing machine learning applications~\cite{das2018dataflow,balaji2019ISVLSIframework,hong2017hierarchical,chen2017using,bacis2017pipelined,shihao_designflow}. However, none of these approaches address application analysis with limited hardware resources, both at design-time and at run-time.

%% file: sections/conclusions.tex
We introduce \tech{} for predictable synthesis of SNN-based applications on state-of-the-art neuromorphic hardware. Prior works have only addressed design-time mapping, considering unlimited resources in the underlying hardware. These approaches present significant limitations when used to compile and map machine-learning applications to a resource-constrained hardware. \tech{} makes five key contributions. 
First, we present an approach to analyze machine-learning programs and generate SNN workload using representative data. 
Second, we present an approach to decompose and partition complex SNN workloads to generate clusters of neurons and synapses such that each cluster can fit onto a crossbar of the hardware. 
Third, we exploit the rich semantics of Synchronous Dataflow Graphs (SDFGs) to represent clustered SNN programs. This allows for the SNN's performance, e.g., throughput, to be estimated on the hardware as a function of key properties such as number of crossbars, dimension of crossbars, buffer space on tiles, and tile communication bandwidth.
Four, we develop a novel scheduling algorithm based on Self-Timed Execution for executing clusters on crossbars of a neuromorphic hardware, providing performance guarantee in scenarios with dynamic resource availability.
Five, we propose a design-space exploration framework incorporating \tech{} that allows the Pareto-space of different SNN mappings to hardware to be explored while considering other hardware metrics such as energy, latency, and reliability.

We evaluate \tech{} using 10 machine learning programs that are representative of the three most commonly used neural network classes --- convolutional neural network (CNN), multi-layer perceptron (MLP), and recurrent neural network (RNN).
Our results demonstrate that \tech{} provides much tighter performance guarantee compared to
%and reduction in compilation time 
current practices.